\documentclass[preprint,3p]{elsarticle}

\input{epsf}

\usepackage{bold-extra}
\usepackage[english]{babel}
\usepackage[latin1]{inputenc}
\usepackage{natbib}
\usepackage{amssymb}
\usepackage{amsmath}  

\usepackage{color}    
\usepackage{moreverb} 
\usepackage{rotating}
\usepackage{tabularx}
\usepackage{stmaryrd}

\usepackage{pdfpages}
\usepackage{morefloats}
\usepackage{graphics}
\usepackage{bbm}
\usepackage{epsfig}
\usepackage{mathrsfs}
\usepackage{multirow}
\usepackage{amsthm}
\usepackage{graphicx} 

\newtheorem{df}{Definition}
\newtheorem{thm}{Theorem}

\newproof{pf}{Proof}

\newcommand{\abs}[1]{\mid #1 \mid}
\newcommand{\sentences}[1]{\left\llbracket #1 \right\rrbracket}

\newcommand{\sentencesw}[1]{\llceil #1 \rrceil}
\newcommand{\ceil}[1]{\left\lceil #1 \right\rceil}
\newcommand{\absd}[1]{\left\Vert #1 \right\Vert}

\newcommand{\dinfty}[0]{\pmb{\pmb{\infty}}}

\journal{Computer Speech and Language}

\begin{document}

\begin{frontmatter}

\title{Adaptive Scheduling for Adaptive Sampling in 
  {\sc pos} Taggers Construction}




\author[UVigo]{Manuel Vilares Ferro\corref{cor}}
\cortext[cor]{Corresponding author: tel. +34 988 387280, fax +34 988 387001.}
\ead{vilares@uvigo.es}
\author[UVigo]{V\'ictor M. Darriba Bilbao}
\ead{darriba@uvigo.es}
\author[UDC]{Jes\'us Vilares Ferro}
\ead{jvilares@udc.es}
\address[UVigo]{Department of Computer Science, University of Vigo \\ Campus As Lagoas s/n, 32004 -- Ourense, Spain}
\address[UDC]{Department of Computer Science, University of A Coru\~na \\ Campus de Elvi\~na, 15071 -- A Coru\~na, Spain}

\begin{abstract}
We introduce an adaptive scheduling for adaptive sampling as a novel
way of machine learning in the construction of part-of-speech
taggers. The goal is to speed up the training on large data sets,
without significant loss of performance with regard to an optimal
configuration. In contrast to previous methods using a random, fixed
or regularly rising spacing between the instances, ours analyzes the
shape of the learning curve geometrically in conjunction with a
functional model to increase or decrease it at any time. The algorithm
proves to be formally correct regarding our working
hypotheses. Namely, given a case, the following one is the nearest
ensuring a net gain of learning ability from the former, it being
possible to modulate the level of requirement for this condition. We
also improve the robustness of sampling by paying greater attention to
those regions of the training data base subject to a temporary
inflation in performance, thus preventing the learning from stopping
prematurely.

The proposal has been evaluated on the basis of its reliability to
identify the convergence of models, corroborating our expectations. While a
concrete halting condition is used for testing, users can choose any
condition whatsoever to suit their own specific needs.
\end{abstract}

\begin{keyword}
correctness \sep learning curve \sep {\sc pos} tagging \sep
robustness \sep sampling scheduling
\end{keyword}

\end{frontmatter}

\section{Introduction}

The possibility of accessing massive amounts of data and the decline
in the cost of disk storage have decisively contributed to the growing
popularity of \textit{machine learning} ({\sc ml}) algorithms as the
basis for modelling tasks in both the
classification~\citep{Leite:2012:SCA:2358856.2358868} and
clustering~\citep{Meek:2002:LSM:944790.944798} domains. However,
managing large amounts of information is an expensive, time-consuming
and non-trivial activity, especially when expert knowledge is
needed. Furthermore, having access to vast data bases does not imply
that {\sc ml} algorithms must use them all and a subset is therefore
preferred, provided it does not reduce the quality of the mined
knowledge. Such observations then supply the same learning power with
far less computational cost and allow the training process to be
speeded up, whilst their nature and optimal size are rarely
obvious. This justifies the interest of developing efficient sampling
techniques, which involves anticipating the link between performance
and experience regarding the \textit{accuracy} of the system we are
generating. At this point, \textit{correctness} with respect to the
working hypotheses and \textit{robustness} against changes to them
should be guaranteed in order to supply a practical solution. The
former ensures the effectiveness of the proposed strategy in the
framework considered, while the latter enables fluctuations in the
learning conditions to be assimilated without compromising
correctness, thus providing reliability to our calculations.

An area of work that is particularly sensitive to these inconveniences
is \textit{natural language processing} ({\sc nlp}), the components of
which are increasingly based on {\sc
  ml}~\citep{Biemann:2006:UPT:1557856.1557859,KATRIN08.335}. This is
due to the major effort required to create the labelled data sets used
as a training basis to generate such tools, especially when they
involve new application domains where resources are scarce or even
non-existent. The problem is especially delicate in the case of
\textit{part-of-speech} ({\sc pos}) \textit{tagging}, the
classification task that marks a word in a text (corpus) as
corresponding to a particular {\sc pos}\footnote{A {\sc pos} is a
  category of words which have similar grammatical properties. Words
  that are assigned to the same {\sc pos} generally display similar
  behaviour in terms of syntax, i.e. they play analogous roles
  within the grammatical structure of sentences. The same applies in
  terms of morphology, in that they undergo inflection for similar
  properties. Commonly listed English {\sc pos} labels are noun, verb,
  adjective, adverb, pronoun, preposition, conjunction, interjection,
  and also numeral, article or determiner.}, based on both its
definition and its context. One reason for this is the complexity of
both the annotation task and the relations to be captured from
learning, but another is that it serves as a first step for other {\sc
  nlp} functionalities such as parsing and semantic analysis, so
errors at this stage can lower their
performance~\citep{Song:2012:CSP:2390524.2390661}. All this makes up a
popular experimentation field for introducing new {\sc ml} facilities,
particularly around sampling
technology~\citep{Bloodgood:2009:MSA:1596374.1596384,Lewis:1994:SAT:188490.188495,Reichart:2010:TLC:1870568.1870579,Schmid:2008:ECP:1599081.1599179,Vlachos:2008:SCA:1349893.1350099},
as with the present work. In this context, we first examine in
Section~\ref{section-state-of-the-art} the methodologies serving as
inspiration to solve the question posed, as well as our
contributions. Next, Section~\ref{section-formal-framework} reviews
the mathematical basis necessary to support our proposal, which we
present in Section~\ref{section-abstract-model}. In
Section~\ref{section-testing-frame}, we describe the testing frame for
the experiments illustrated in
Section~\ref{section-experiments}. Finally,
Section~\ref{section-conclusions} presents our final conclusions.


\section{The state of the art}
\label{section-state-of-the-art}

While the common goal is to choose a set of observations and determine
whether it is large enough to reach the desirable learning
performance, we characterize a sampling strategy according to three
labels often compatible with each other. In light of the consideration
or non-consideration of an expert opinion for selecting the sample,
whether human or not, the algorithm is
identified~\citep{Cohn:1994:IGA:189256.189489,Saar-Tsechansky04} as
\textit{active} or \textit{non-active}. Depending on the use or
non-use of knowledge about the behaviour of the model to be generated,
we distinguish~\citep{John96staticversus} between \textit{dynamic} and
\textit{static sampling}. We can finally
differentiate~\citep{Chen:2013:PNA:2568488.2568754} between
\textit{adaptive}, also called
\textit{sequential}~\citep{Domingo:2002:ASM:593433.593526} or
\textit{progressive}~\citep{Watanabe:2005:SST:1162426.1162428}, and
\textit{batch sampling} when the size of the sample is determined
iteratively in an online fashion or \textit{prior to} commencing the
selection task. In order to mark the end of the sequencing process,
adaptive methods associate a \textit{halting condition}. At all
events, although these labels are not mutually exclusive, the
scheduling strategy applied to a great extent conditions the sampling
approach.

\subsection{Sampling scheduling}

Active sampling often associates an adaptive architecture, recruiting
for annotation at each cycle only examples corresponding to near miss
observations~\citep{Winston75}, i.e. negative ones that differ from
the learned concept in a small number of significant points. This
results in a two-stage procedure in which a reduced set of labelled
cases is first collected to start a loop of selection and further
reprocessing on the complete training data base until a stopping
condition verifies. Most of this research focuses on
\textit{pool-based active sampling}, in which selection is made from a
pool following two main schema:
\textit{uncertainty}~\citep{Lewis:1994:SAT:188490.188495} and
\textit{query-by-committee}~\citep{Seung:1992:QC:130385.130417}. The
former uses a single classifier to select the observation on which it
has the lowest certainty. Committee-based sampling converges to the
optimal model more quickly~\citep{Freund:1997:SSU:263100.263123} by
considering a set of classifiers working on the principle of maximal
disagreement among them. Unfortunately, active sampling is
windowing~\citep{Quinlan83} so its learning curves are notoriously
ill-behaved on noisy data~\citep{Provost:1999:EPS:312129.312188},
increasing the amount of such random fluctuations on subsequent
samples. Accordingly, performance often decreases as the process
progresses~\citep{Furnkranz98} questioning, despite its apparent
potential, its
adoption~\citep{Attenberg:2011:ILD:1964897.1964906}. For that reason
we do not cover it in this work.

Focusing on non-active designs, static proposals determine the size of
the sample from its representativeness of the training data base in
terms of feature distributions. This can be done through
\textit{simple} procedures such as consideration of the complete set
when the cost is affordable, a technique commonly known as
\textit{trivial selection}, or a part of this suggested by an
omniscient oracle. The \textit{random selection} of a fixed number or
fraction of observations can also be considered. All of these are
batch techniques for which, with the exception of the trivial
approach, no formal interpretation for their correctness is
possible. This justifies the recourse to adaptive methods, where the
simplest and most common way to pick instances is again
randomly~\citep{AounallahQuirionMineau04}. Given an initial sample
size and a schedule of sample size increments, new instances are then
added until the distributions of both the sample and the training data
base, are sufficiently similar. Alternatively, a fixed sequencing
scheduling can be considered, typically by applying
\textit{geometric}~\citep{Provost:1999:EPS:312129.312188} or
\textit{arithmetic selection}~\citep{John96staticversus}, also
referred to as \textit{uniform selection}~\citep{WeissTian08}.  In
either case, static sampling uses statistical
inference~\citep{CaseBerg:01} to define the halting condition and, in
fact, practitioners sometimes speak of \textit{statistically valid
  sampling} to refer to it~\citep{John96staticversus}. This makes it
possible to derive a theoretically guaranteed sample size, sufficient
to achieve a task with given confidence by using the so-called
\textit{concentration bounds}~\citep{Chernoff52,Hoeffding63}, which
provides a well-founded basis to introduce these kinds of
techniques. So, its correctness and robustness have been formally
demonstrated~\citep{Domingo:2002:ASM:593433.593526}, its computational
complexity analyzed~\citep{Lynch2003}, and its usefulness for scaling
up learning algorithms in data mining applications
proved~\citep{Watanabe:2005:SST:1162426.1162428}. Sadly, the number of
instances can be overestimated or even unrealistic~\citep{Lipton1993},
making the static option less attractive.

All the above justifies the interest in the more flexible dynamic
sampling. It is then possible to work guided by a model for the
shape of the learning curve~\citep{KadiePhd}, which we assume slows to
an almost horizontal slope at about the time when the true performance
reaches its peak. This suggests a sequential scheduling such as those
previously commented, which results in an adaptive philosophy, where at
each cycle a model is built from the current sample and its
performance evaluated. In this regard, arithmetic progressions can
require an unreasonable number of iterations when a large number of
cases is needed. In contrast, a geometric schedule quickly reach an
appropriate sample size, while it may easily overfit local disruptions
and thus stop ahead of time due to a momentary increase in
performance~\citep{Last:2009:IDM:1557019.1557076}, resulting in a
fragil robustness.

Regarding the halting condition, we envisage two approaches in
accordance with the consideration of predictive accuracy as an
absolute stopping criterion~\citep{FreyFisher99} or as nothing more
than a cost factor of an optimization problem stated in
\textit{decision theory}~\citep{Howard66}. The first scenario involves
identifying the final plateau of the learning curve in terms of
functional convergence. Some procedures in this respect have become
popular, such as \textit{local detection} and \textit{learning curve
  estimation}~\citep{John96staticversus}, or \textit{linear regression
  with local sampling}~\citep{Provost:1999:EPS:312129.312188}, even if
we have had to wait until recently~\citep{VilaresDarribaRibadas16} to
dispose of a formally correct one. On the contrary, when sampling
performance is understood as the search for a proper cost/benefit
trade-off, the authors have recourse to statistically based
strategies. Formally, they apply the principle of \textit{maximum
  expected utility} ({\sc
  meu})~\citep{Meek:2002:LSM:944790.944798}. This implies taking into
account all effectiveness considerations, which depends on the degree
of control exercised by the user on the learning process. In its
absence, i.e. using non-active techniques as we do, the final cost is
the sum of data acquisition, error and model induction
charges~\citep{WeissTian08}. Nonetheless, at best, heuristic
techniques are used to calculate the first two and there is thus no
way of guaranteeing the location of a global
optimum~\citep{Last:2009:IDM:1557019.1557076}, which often results in
assuming fixed budgets~\citep{Kapoor05}. That is why this kind of
stopping criterion is not advisable to define reliable testing frames
on sampling scheduling.

In practice, the success of adaptive sampling depends heavily on prior
knowledge about the underlying model induction algorithm applied,
which may be less than precise, thereby precipitating or delaying the
detection of convergence and increasing the associated operating
costs. Since this expertise can be obtained on the fly, the use of
also adaptive scheduling seems to be better placed to achieve optimal
results. It therefore becomes a question of rationally adjusting the
number of cases between successive cycles. Surprisingly, to the best
of our knowledge, the only action in this direction is due to Provost
\textit{et al.}~\cite{Provost:1999:EPS:312129.312188}. From a set of
instances large enough to obtain reliable estimates, they iteratively
build models for both the convergence probability distribution and the
run-time complexity of the underlying induction algorithm. At each
cycle the convergence is checked and, if it does not occur, the
schedule is rebuilt from the latest information and the process
restarts. However, this proposal performs in practice much worse than
the less complicated geometric one, which seems to have discouraged
further research on this topic, despite its vast potential.

\subsection{Our contribution}

We introduce an adaptive scheduling, baptized as {\sc
  colts}\footnote{After {\sc co}ncavity {\sc l}imi{\sc t} {\sc
    s}cheduling.}, with a view to reducing operating charges in
non-active adaptive sampling. The idea is to calculate, at each
iteration, the smallest amount of training data to be added for
ensuring that the next case is relevant in learning terms. From a set
of usual working hypotheses in {\sc ml}, the technique is described
taking an exclusively geometrical point of view. Once a sequence of
observations has been set, a functional approximation to the
associated partial learning curve is built. The next instance from
which a new observation to update our evaluation is mandatory,
i.e. from which the working hypotheses can no longer be
guaranteed, is then located. We do this by calculating the case the
degree of concavity, namely the learning speed, which cannot be
maintained over time on the real learning curve. The correctness of
the method is formally established and its robustness explored.

The proposal is evaluated within a uniform testing framework, in the
sense that its standards of evidence do not favour any particular
sampling scheduling, taking the generation of {\sc pos} taggers as a
case study. Once a learner, a halting condition and a training data
base are fixed, the aim is to categorize a set of schedules according
to their efficiency to achieve a given level of accuracy in the model
being generated. To that end, predictive accuracy is taken as the
stopping criterion, thereby avoiding the inconveniences associated
with {\sc meu}-based halting conditions. We also introduce the metric
used as an assessment basis, together with its associated monitoring
architecture for data collection. The latter captures the concept of
testing round (\textit{run}), which serves to normalize the conditions
under which the experiments take place. Thus, runs only
distinguishable by their scheduling strategy are grouped around an
item acting as baseline, in what we call a \textit{local testing
  frame}. It then becomes possible to compare, within these
structures, runs in terms of both training resources used and overall
learning costs. By doing so, we avoid recourse to cumbersome
heuristics, often highly dependent on the knowledge domain considered,
endowing the tests with reliability and safety.

\section{The formal framework}
\label{section-formal-framework}

The aim is to introduce the mathematical basis that enables to prove
the correctness of our proposal. Most of these formal notions are
taken from Vilares \textit{et al.}~\cite{VilaresDarribaRibadas16},
denoting the set of real numbers by $\mathbb{R}$ and that of naturals
by $\mathbb{N}$, assuming that $0 \not\in \mathbb{N}$. Another
preliminary question to be clarified, because the generation of {\sc
  ml}-based {\sc pos} taggers serves as illustration guide, is the
identification of the accuracy concept usually accepted in that kind
of model. We define it as the number of correctly tagged tokens
divided by the total ones, expressed as a
percentage~\citep{VanHalteren1999} and calculated following some
generally admitted usages: all tokens are counted, including
punctuation marks, and it is supposed that only one tag \textit{per}
token is provided.

\subsection{The working hypotheses}

We start with a sequence of observations calculated from cases
incrementally taken from a training data base, meeting some conditions
to ensure a predictable progression of the estimates over a virtually
infinite interval. So, they are assumed to be independently and
identically
distributed~\cite{Domingo:2002:ASM:593433.593526,Schutze:2006:PTP:1183614.1183709,KATRIN08.335}. We
then accept that a learning curve is a positive definite and strictly
increasing function on $\mathbb{N}$, where numbers are the positions of
instances in the training data set, and upper bounded by 100. This
results in a concave graph with horizontal asymptote.

\begin{figure}[htbp]
\begin{center}
\begin{tabular}{cc}
\hspace*{-.5cm}
\includegraphics[width=0.47\textwidth]{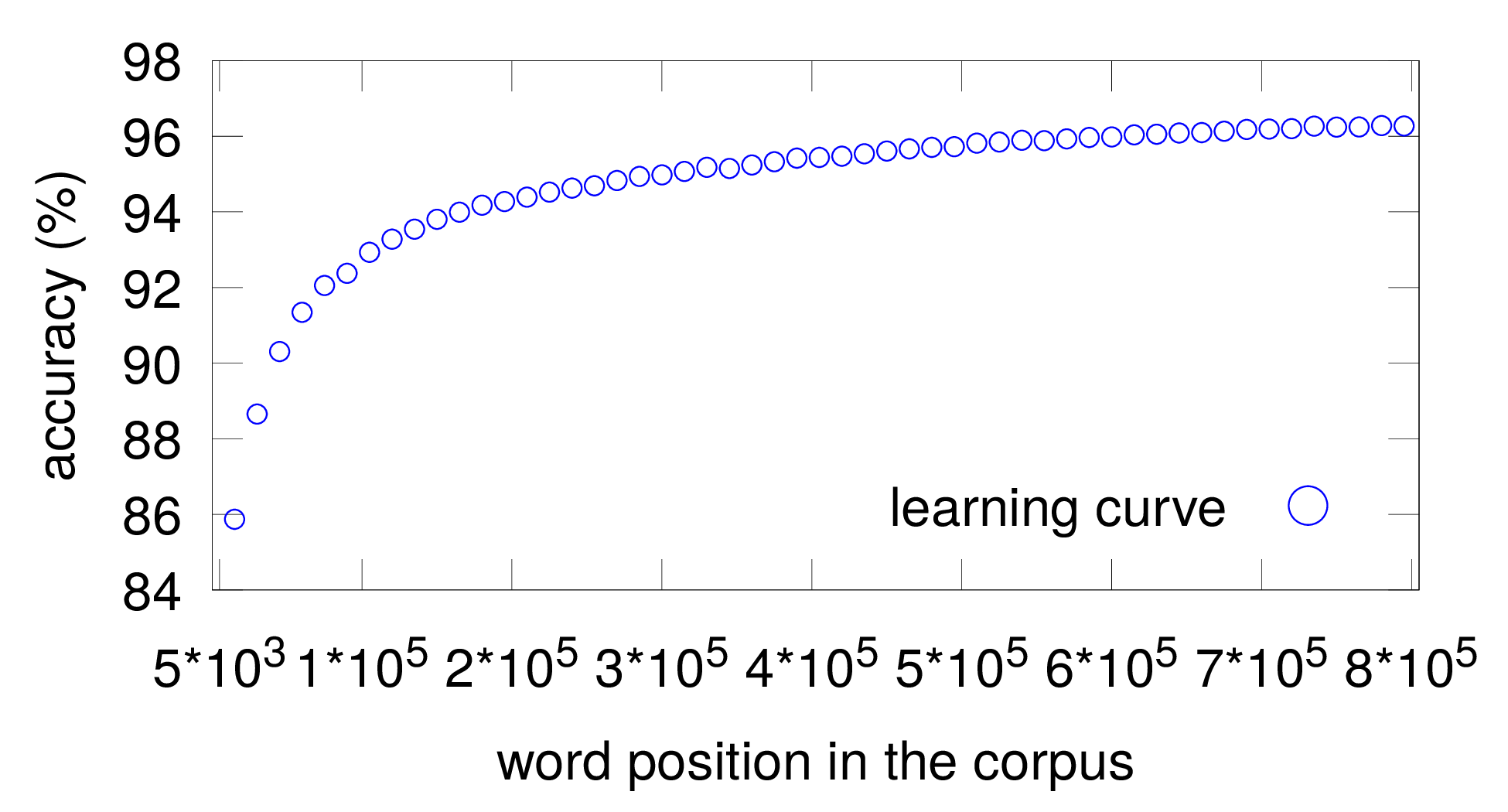}
&
\includegraphics[width=0.47\textwidth]{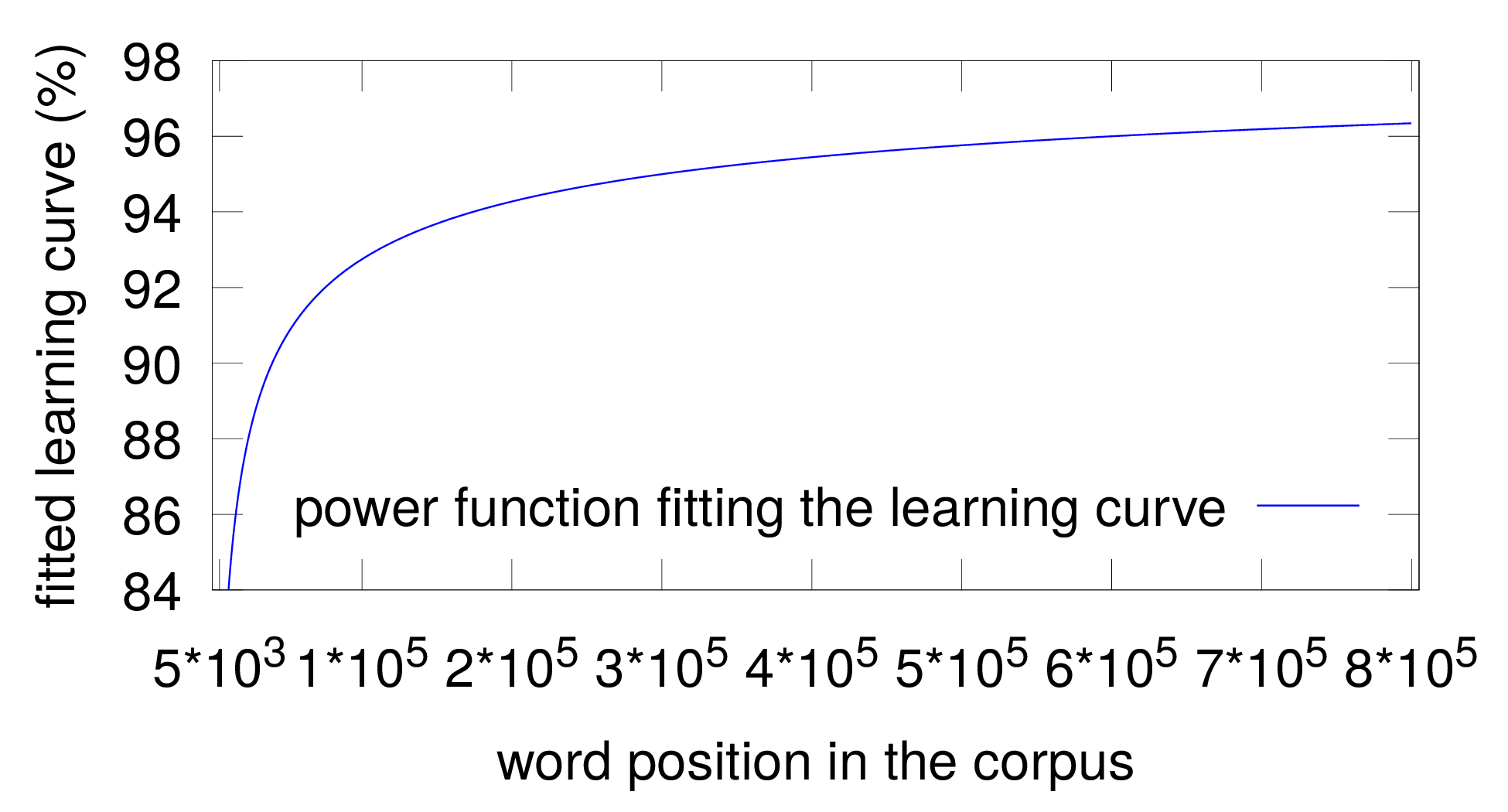}
\end{tabular}
\caption{Learning curve for fn{\sc tbl} on {\sc f}rown, and an
  accuracy pattern fitting it.}
\label{fig-accuracy-fnTBL-Frown-5000-800000}
\end{center}
\end{figure}

Such hypotheses make up an idealized working frame to support
correctness, while real learners may deviate from it, justifying a
later study of robustness. These deviations translate into
irregularities in both concavity and increase of the learning curves,
as shown in the left-most diagram of
Fig.~\ref{fig-accuracy-fnTBL-Frown-5000-800000} for the training of
the {\it fast transformation-based learning} (fn{\sc tbl})
tagger~\citep{Ngai2001} on the \textit{Freiburg-Brown} ({\sc f}rown)
corpus of American English~\citep{Mair2007}. The cases are therefore
indexed by the position of a word in the text.

\subsection{The notational support}

Having identified the working hypotheses, we need to formalize the data
structures we are going to work with, such as the progressive sequence
of instances whose selection we want to optimize.

\begin{df}
\label{def-learning-scheme}
Let ${\mathcal D}$ be a training data base,
$\mathcal K \subsetneq \mathcal D$ a subset of initial items from $D$
and
$\sigma \! \in \! \Sigma \! := \{\zeta:\mathbb{N} \rightarrow
\mathbb{N}\}$ a function. We define a {\em learning scheme} for
$\mathcal D$ with {\em kernel} ${\mathcal K}$ and {\em step} $\sigma$,
as a triple
$\mathcal{D}^{\mathcal{K}}_{\sigma}=[\mathcal{K},\sigma,\{\mathcal
D_i\}_{i \in \mathbb{N}}]$ with $\{\mathcal D_i\}_{i \in \mathbb{N}}$
a cover of $\mathcal{D}$ verifying:
\begin{equation}
{\mathcal D}_1 := {\mathcal K} \mbox{ and }
{\mathcal D}_i := {\mathcal
  D}_{i-1} \cup {\mathcal I}_{i}, \; \mathcal I_i
\subset {\mathcal D} \setminus {\mathcal
  D}_{i-1}, \; \absd{{\mathcal I}_{i}}=\sigma(i), \; \forall i \geq 2
\end{equation}
\noindent where $\absd{{\mathcal I}_{i}}$ is the cardinality of
${\mathcal I}_{i}$. We refer to $\mathcal{D}_i$ as the {\em individual
  of level} $i$ {\em for} $\mathcal{D}^{\mathcal{K}}_{\sigma}$.
\end{df}

A learning scheme relates a level $i$ with the position
$\absd{\mathcal D_i}$ in the training data base, determining the
sequence of observations $\{[x_i, {\mathcal A}_{\dinfty{}}[{\mathcal
      D}](x_i)], \; x_i := \absd{\mathcal D_i} \}_{i \in \mathbb{N}}$,
where ${\mathcal A}_{\dinfty{}}[{\mathcal D}](x_i)$ is the accuracy
achieved on such instance by the learner. Thus, a level determines an
iteration in the adaptive sampling whose learning curve is ${\mathcal
  A}_{\dinfty{}}[{\mathcal D}]$, whilst ${\mathcal K}$ delimits a
portion of ${\mathcal D}$ we believe to be enough to initiate
consistent evaluations of the training. For its part, $\sigma$
identifies the sampling scheduling. As we want to address the latter
from geometrical criteria, we need to extrapolate the partial learning
curves according to a functional pattern providing stability to the
estimates. The focus is on curves that verify the working hypotheses,
but are also infinitely differentiable over the training domain. This
supplies graphs without disruptions due to instantaneous jumps while
ensuring their regularity.

\begin{df}
\label{def-accuracy-pattern-fitting}
Let $C^\infty_{(0,\infty)}$ be the C-infinity functions in
$\mathbb{R}^{+}$, we say that $\pi: \mathbb{R}^{{+}^{n}} \rightarrow
C^\infty_{(0,\infty)}$ is an {\em accuracy pattern} iff $\pi(a_1,
\dots, a_n)$ is positive definite, upper bounded, concave and strictly
increasing.
\end{df}

An example of accuracy pattern is the \textit{power family} of curves
$\pi(a,b,c)(x) :=-a * x^{-b}+c$, hereafter used as running one. Its
upper bound is the horizontal asymptote value \(\lim \limits_{x
  \rightarrow \infty} \pi(a,b,c)(x) = c\), and
\begin{equation}
\pi(a,b,c)'(x)=a * b * x^{-(b+1)} > 0 \hspace*{.75cm} 
\pi(a,b,c)''(x)=-a * b
* (b+1) * x^{-(b+2)} < 0 
\end{equation}
\noindent which guarantees increase and concavity in $\mathbb{R}^{+}$,
respectively. This is illustrated in the right-most diagram of
Fig.~\ref{fig-accuracy-fnTBL-Frown-5000-800000}, whose goal is to fit
the learning curve represented in the left-hand side. Here, the values
$a=542.5451$, $b=0.3838$ and $c=99.2876$ are provided by the
\textit{trust region method}~\citep{Branch1999}, a regression
technique minimizing the summed square of \textit{residuals}, i.e.
the differences between the observed values and the fitted
ones. Furthermore, as it is intended to determine from the current
case the next one ensuring significance for learning, such a
concept of usefulness must be formalized. To do it, we need to
evaluate the progression of accuracy during the training process,
which results in studying the sequence of curves modelled from the
partial learning ones.

\begin{df}
\label{def-trace}
Let $\mathcal{D}^{\mathcal{K}}_{\sigma}$ be a learning scheme, $\pi$
an accuracy pattern and $\ell \in \mathbb{N}, \; \ell \geq 3$ a
position in the training data base $\mathcal{D}$. We define the {\em
  learning trend of level} $\ell$ {\em for} ${\mathcal
  D}^{\mathcal{K}}_{\sigma}$ {\em using} $\pi$, as a curve ${\mathcal
  A}_{\ell}^\pi[{\mathcal D}^{\mathcal{K}}_{\sigma}] \in \pi$, fitting
the observations $\{[x_i, {\mathcal A}_{\dinfty{}}[{\mathcal
      D}](x_i)], \; x_i := \absd{\mathcal D_i}
\}_{i=1}^{\ell}$. A sequence of learning trends ${\mathcal
  A}^\pi[{\mathcal D}^{\mathcal {K}}_{\sigma}] :=\{{\mathcal
  A}_{\ell}^\pi[{\mathcal D}^{\mathcal{K}}_{\sigma}]\}_{\ell \in
  \mathbb{N}}$ is called a {\em learning trace}. We refer to
$\{\alpha_\ell\}_{\ell \in \mathbb{N}}$ as the {\em asymptotic
  backbone} of ${\mathcal A}^\pi[{\mathcal D}^{\mathcal
    {K}}_{\sigma}]$, where $y = \alpha_\ell := \lim \limits_{x
  \rightarrow \infty} {\mathcal A}_{\ell}^\pi[{\mathcal
    D}^{\mathcal{K}}_{\sigma}](x)$ is the asymptote of ${\mathcal
  A}_\ell^\pi[{\mathcal D}^{\mathcal {K}}_{\sigma}]$.
\end{df}

A learning trend
${\mathcal A}_{\ell}^\pi[{\mathcal D}^{\mathcal{K}}_{\sigma}]$
requires a level $\ell \geq 3$, because we need at least three
observations to generate a curve. Its value
${\mathcal A}_{\ell}^\pi[{\mathcal D}^{\mathcal{K}}_{\sigma}](x_i)$
represents the prediction for accuracy on a case $x_i$, using a model
generated from the first $\ell$ iterations of the
learner. Accordingly, the asymptotic term $\alpha_\ell$ is nothing
other than the estimate for the highest accuracy attainable. This way,
a learning trace gives a comprehensive picture of the increase in
accuracy due to new observations, as well as future expectations in
that respect. Continuing with the tagger fn{\sc tbl} and the corpus
{\sc f}rown, Fig. 2 shows (left) a portion of the learning trace
with kernel and uniform step function $5*10^3$, also including the
real learning curve and a zoom view (right). As our running frame is
the generation of {\sc pos} taggers, levels are hereafter indicated by
word positions in the training corpus. At this point, we are ready to
capture the notion of learning utility associated to a case.

\begin{figure}[htbp]
\begin{center}
\begin{tabular}{cc}
\hspace*{-.45cm}
\includegraphics[width=0.47\textwidth]{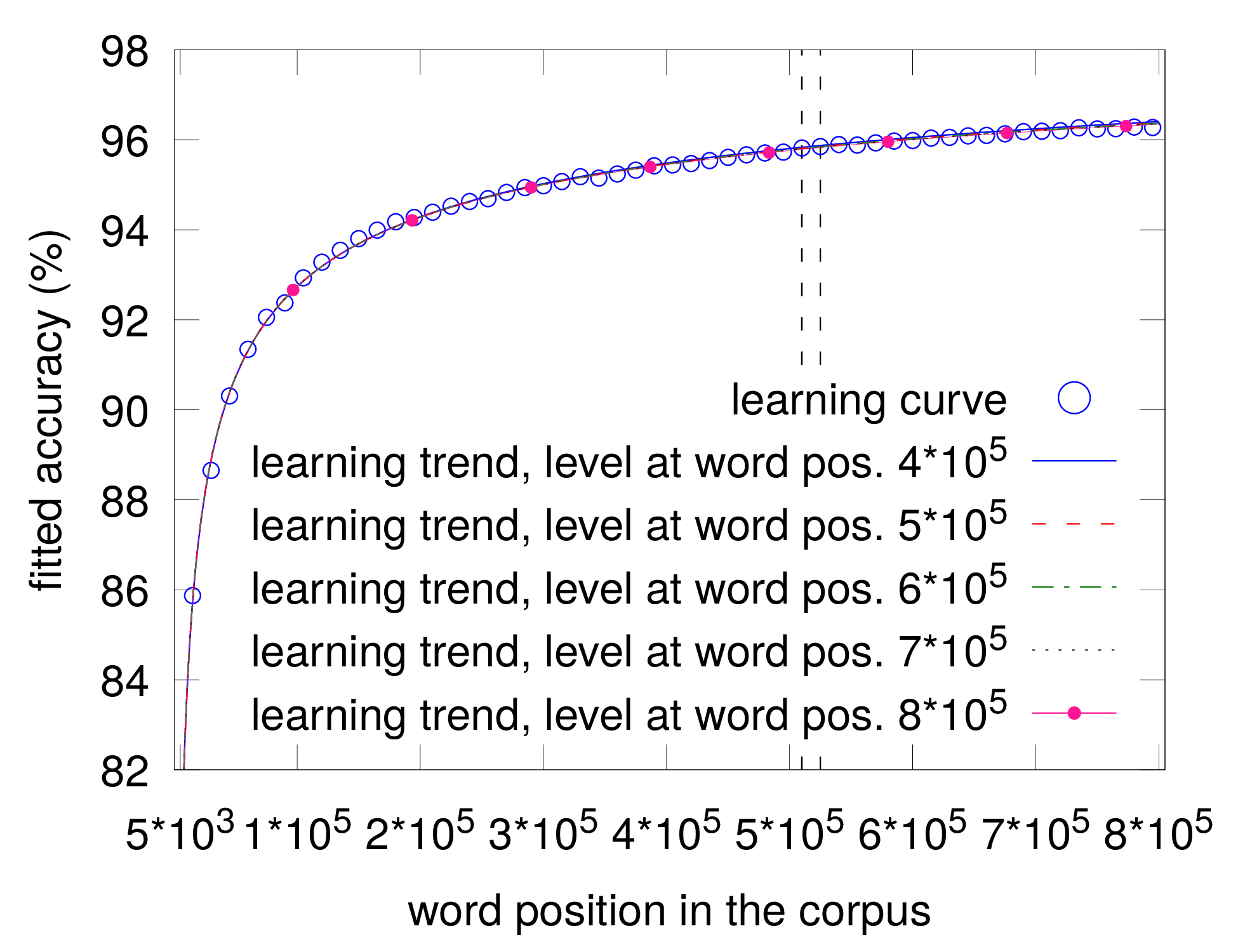}
&
\includegraphics[width=0.47\textwidth]{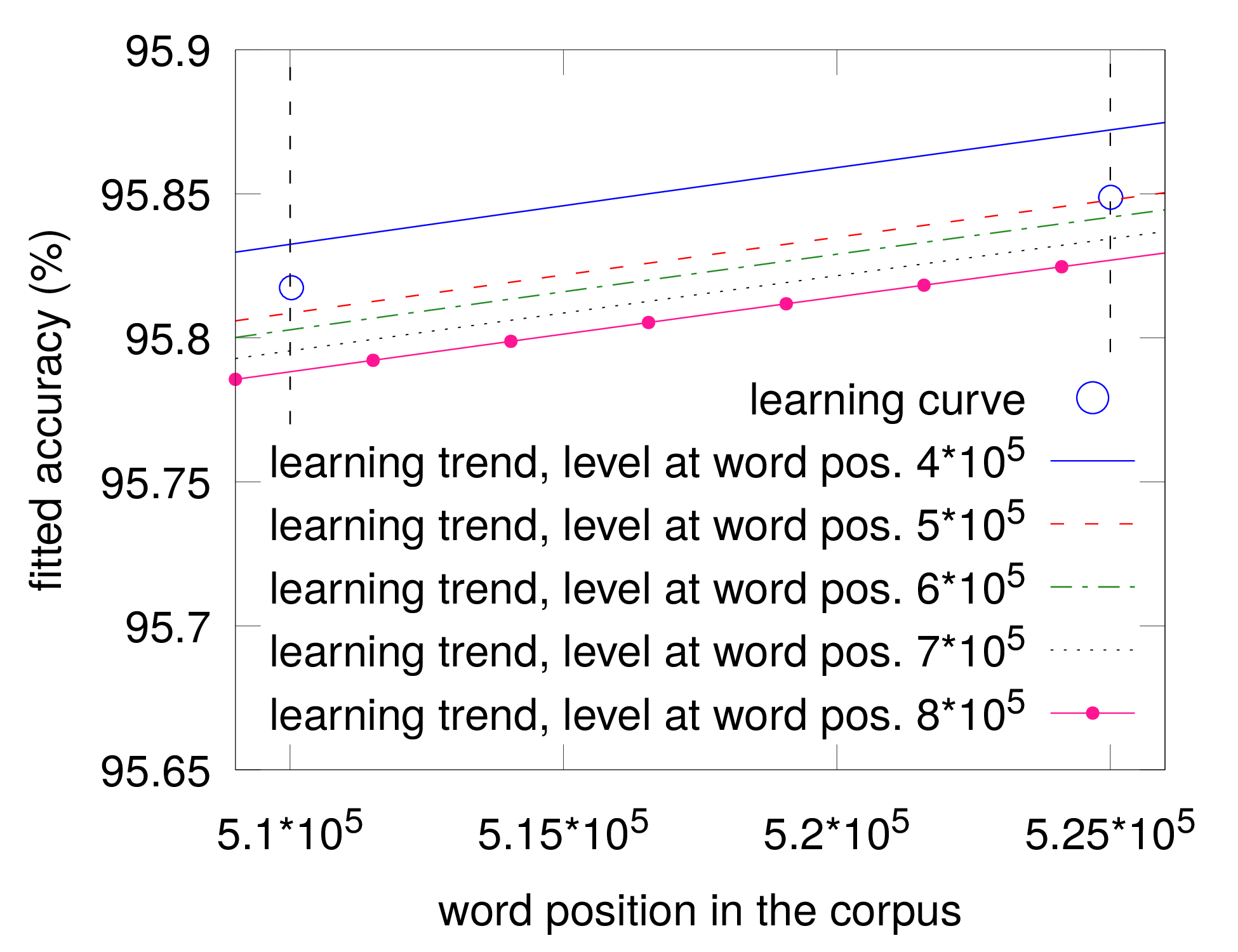}
\end{tabular}
\caption{Learning trace for fn{\sc tbl} on {\sc f}rown.}
\label{fig-trace-and-level-sequence-nlls-fnTBL-Frown-5000-800000}
\end{center}
\end{figure}

\begin{df}
\label{def-relevant-instance}
Let ${\mathcal A}^\pi[{\mathcal D}^{\mathcal {K}}_{\sigma}]$ be a
learning trace and $\ell \in \mathbb{N}, \; \ell \geq 3$. We say that
$x_{\ell+1}$ {\em is relevant for} ${\mathcal A}^\pi[{\mathcal
    D}^{\mathcal {K}}_{\sigma}]$ iff ${\mathcal A}_\ell^\pi[{\mathcal
    D}^{\mathcal {K}}_{\sigma}]'(x_\ell) \neq {\mathcal
  A}_{\ell+1}^\pi[{\mathcal D}^{\mathcal {K}}_{\sigma}]'(x_{\ell+1})$,
with $x_{\ell} := \absd{\mathcal D_{\ell}}$.
\end{df}

A case is relevant when the slope of its learning trend at that point
varies with regard to what happens in the previous instance. This
reveals a change in the learning speed, i.e. in the degree of
concavity, observed on the learning curve we try to approximate. We
are therefore talking about an effective step towards the
identification of convergence for training, providing a practical
sense to the notion of relevance.

\section{The abstract model}
\label{section-abstract-model}

We lay the theoretical foundations of our proposal to later interpret
them from an operational point of view. The first objective is to
establish its correctness, i.e. to formalize a sampling scheduling
for which the distance between two consecutive cases is the shortest
one guaranteeing the relevance of the most recent instance. All that
is required for such a purpose is to state the adequate step function.

\subsection{Correctness}

Given ${\mathcal D}$ a training data base, the goal is to identify the
step function $\sigma$ under the terms outlined above, taking into
account that excessively short steps can unnecessarily overload the
sampling procedure, as with those regions in $\mathcal{D}$ on which
the slope of the learning curve does not vary much.

\begin{thm}
\label{th-adaptive-step-size}
Let ${\mathcal A}^\pi[{\mathcal D}^{\mathcal {K}}_{\sigma}]$ be a
learning trace, then:
\begin{equation}
\forall i \geq 3, \; \sigma(i+1) \geq 
\frac{\alpha_i - 
      {\mathcal A}_i^\pi[{\mathcal D}^{\mathcal {K}}_{\sigma}](x_i)}
      {{\mathcal A}_i^\pi[{\mathcal D}^{\mathcal {K}}_{\sigma}]'(x_i)}
\Rightarrow 
{\mathcal A}_{i+1}^\pi[{\mathcal D}^{\mathcal {K}}_{\sigma}]'(x_{i+1})
\neq
{\mathcal A}_i^\pi[{\mathcal D}^{\mathcal {K}}_{\sigma}]'(x_i)
\end{equation}
\noindent with $x_i := \absd{\mathcal D_i}$, and $y=\alpha_i$
the horizontal asymptote for ${\mathcal A}_i^\pi[{\mathcal
    D}^{\mathcal {K}}_{\sigma}]$.
\end{thm}

\begin{figure}[htbp]
\begin{center}
\includegraphics[width=0.75\textwidth]{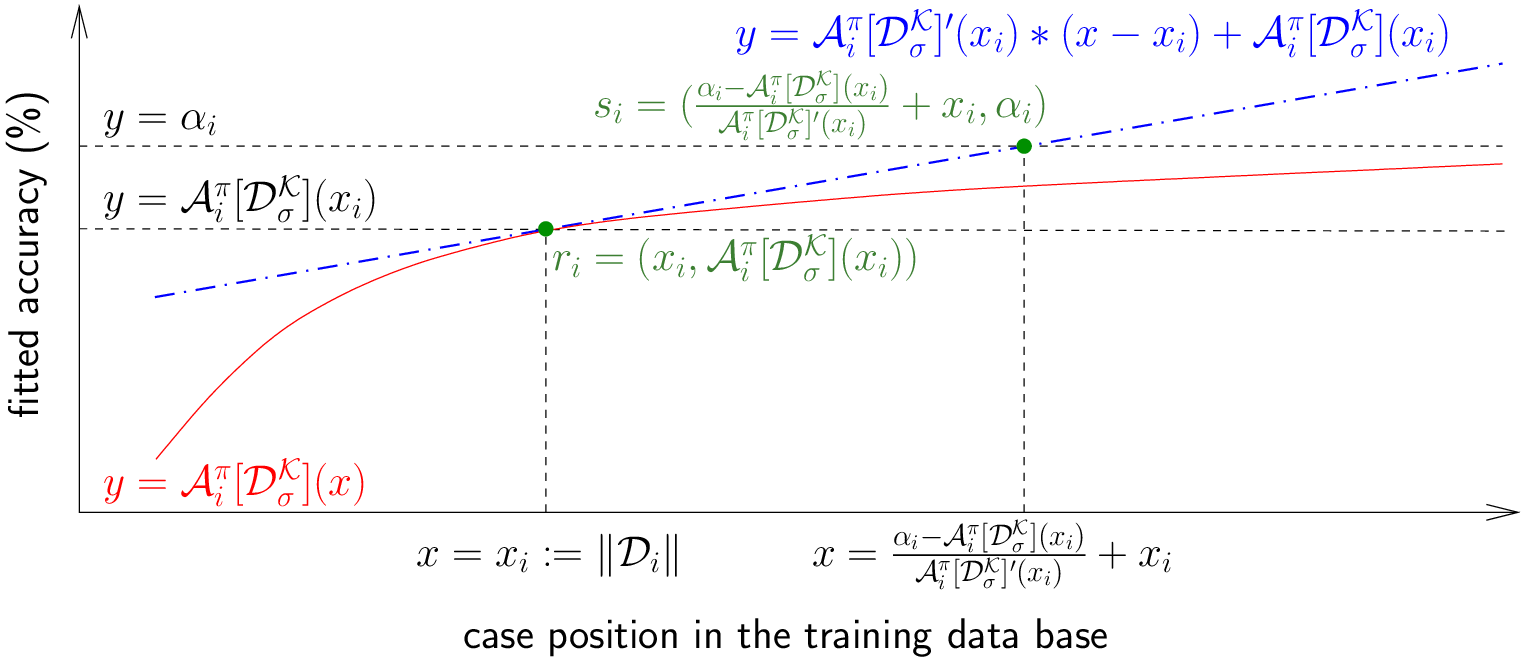}
\end{center}
\caption{Computing dynamically the size of individuals in a learning trace.}
\label{fig-adaptative-step-size}
\end{figure}

\begin{pf}
Suppose a learning trend ${\mathcal A}_i^\pi[{\mathcal D}^{\mathcal
    {K}}_{\sigma}]$ with $i \geq 3$, as shown in
Fig.~\ref{fig-adaptative-step-size}. Given that it is monotonic
increasing (resp. concave), no point on it has an ordinate
(resp. slope) greater than $\alpha_i$ (resp. ${\mathcal
  A}_i^\pi[{\mathcal D}^{\mathcal {K}}_{\sigma}]'(x_i)$) in the
interval $(0, \infty)$ (resp. $(x_i,\infty)$). Accordingly, the slope
of ${\mathcal A}_i^\pi[{\mathcal D}^{\mathcal {K}}_{\sigma}]$ on $x_i$
cannot be maintained beyond the point indicated by the abscissa of
$s_i$, the intersection point between its tangent line thorough $r_i$
and its horizontal asymptote. Since this abscissa is calculated
substituting $y=\alpha_i$ in the tangent
\begin{equation}
y = {\mathcal A}_i^\pi[{\mathcal D}^{\mathcal {K}}_{\sigma}]'(x_i)
\ast (x - x_i) + {\mathcal A}_i^\pi[{\mathcal D}^{\mathcal
    {K}}_{\sigma}](x_i)
\end{equation}
\noindent we have that $s_i = (\frac{\alpha_i - {\mathcal
    A}_i^\pi[{\mathcal D}^{\mathcal {K}}_{\sigma}](x_i)} {{\mathcal
    A}_i^\pi[{\mathcal D}^{\mathcal {K}}_{\sigma}]'(x_i)} + x_i,
\alpha_i)$, from which we conclude the thesis.  $\blacksquare$
\end{pf}

We now have a sufficient condition to identify at each sampling cycle
the closest instance from which a real impact on learning capacity is
ensured, because the step applied guarantees its relevance. However,
this does not exclude the possibility that smaller steps could
eventually produce the same effect. Thus, the probability of a case
being relevant is proportional to the value proposed, an idea that is
useful to formalize.

\begin{df}
Let ${\mathcal A}^\pi[{\mathcal D}^{\mathcal {K}}_{\sigma}]$ be a
learning trace, its {\em probability of relevant training
({\sc port}) at level $i \geq 4$} is 
\begin{equation}
\label{eq-PORT}
\varrho(i) :=
\left\{\begin{array}{ll}
       1 & \mbox{\em if} \; \sigma(i) \geq \mu(i) \\
       \sigma(i)/\mu(i) & \mbox{\em otherwise}
       \end{array} 
\right., \; \mu(i) := \frac{\alpha_i - {\mathcal
    A}_i^\pi[{\mathcal D}^{\mathcal
      {K}}_{\sigma}](x_i)}{{\mathcal
    A}_i^\pi[{\mathcal D}^{\mathcal
             {K}}_{\sigma}]'(x_i)}, \; x_i := \absd{{\mathcal D}_{i}}
\end{equation}
\end{df}

Following Theorem~\ref{th-adaptive-step-size}, $\mu(i)$ is the
shortest separation that guarantees the relevance of the case at level
$i$. Consequently, any step $\sigma(i) \geq \mu(i)$ corresponds to a
maximal {\sc port}, whereas the low distances are associated with
proportional values. Since step functions are strictly positive
definite, the {\sc port} is defined in the interval $(0,1]$ and
  provides a simple mechanism to regulate, in probabilistic terms, the
  interrelation between the speed of learning and the training
  sequence in a learning trace. This allows us to immediately prove
  the correctness of a sample with respect to a given {\sc port}.

\begin{thm}
\label{th-step-function-PORT}
{\em (Correctness)} Let ${\mathcal A}^\pi[{\mathcal D}^{\mathcal
    {K}}_{\sigma[\varrho]}]$ be a learning trace and $\sigma[\varrho]$
the step function 
\begin{equation}
\label{eq-step-function-PORT}
\sigma[\varrho](i) := \ceil{\varrho \ast \frac{\alpha_i - {\mathcal
    A}_i^\pi[{\mathcal D}^{\mathcal
      {K}}_{\sigma[\varrho]}](x_i)} {{\mathcal
    A}_i^\pi[{\mathcal D}^{\mathcal
      {K}}_{\sigma[\varrho]}]'(x_i)}}, \; \varrho
\in (0,1], \; x_i := \absd{{\mathcal D}_{i}}, \; \forall i \geq 4
\end{equation}
\noindent with $y=\alpha_i$ the horizontal asymptote for ${\mathcal
  A}_i^\pi[{\mathcal D}^{\mathcal {K}}_{\sigma[\varrho]}]$ and
$\ceil{x}$ the {\em ceiling function} mapping $x \in \mathbb{R}$ to the
supremum in $[x, \infty) \cap \mathbb{N}$. Then, ${\mathcal
    A}^\pi[{\mathcal D}^{\mathcal {K}}_{\sigma[\varrho]}]$ has the
  smallest {\sc port} greater than or equal to $\varrho$.
\end{thm}

\begin{pf}
Trivial from Theorem~\ref{th-adaptive-step-size}. $\blacksquare$
\end{pf}

\begin{figure}[htbp]
\begin{center}
\begin{tabular}{cc}
\hspace*{-.5cm}
\includegraphics[width=0.47\textwidth]{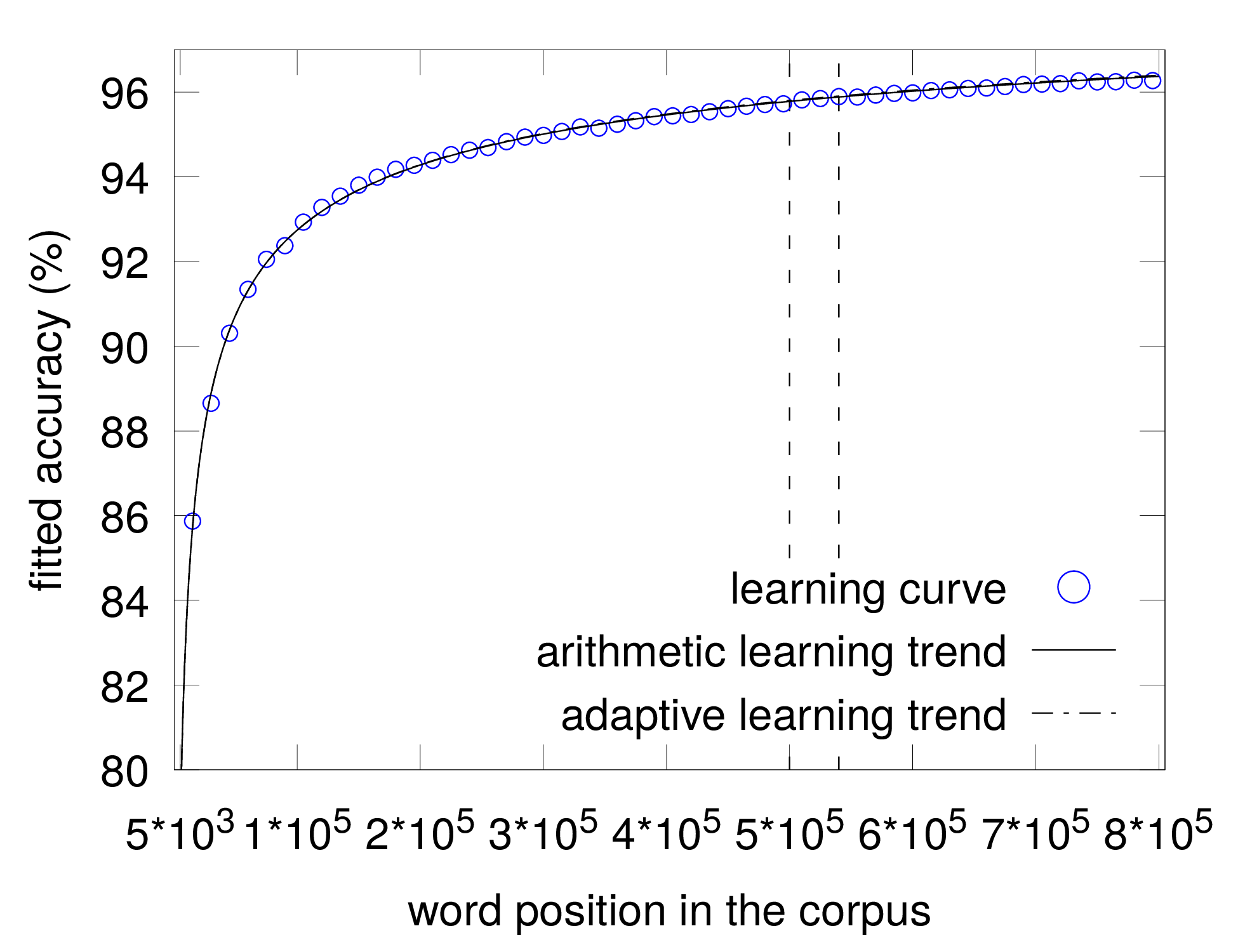}
& 
\includegraphics[width=0.47\textwidth]{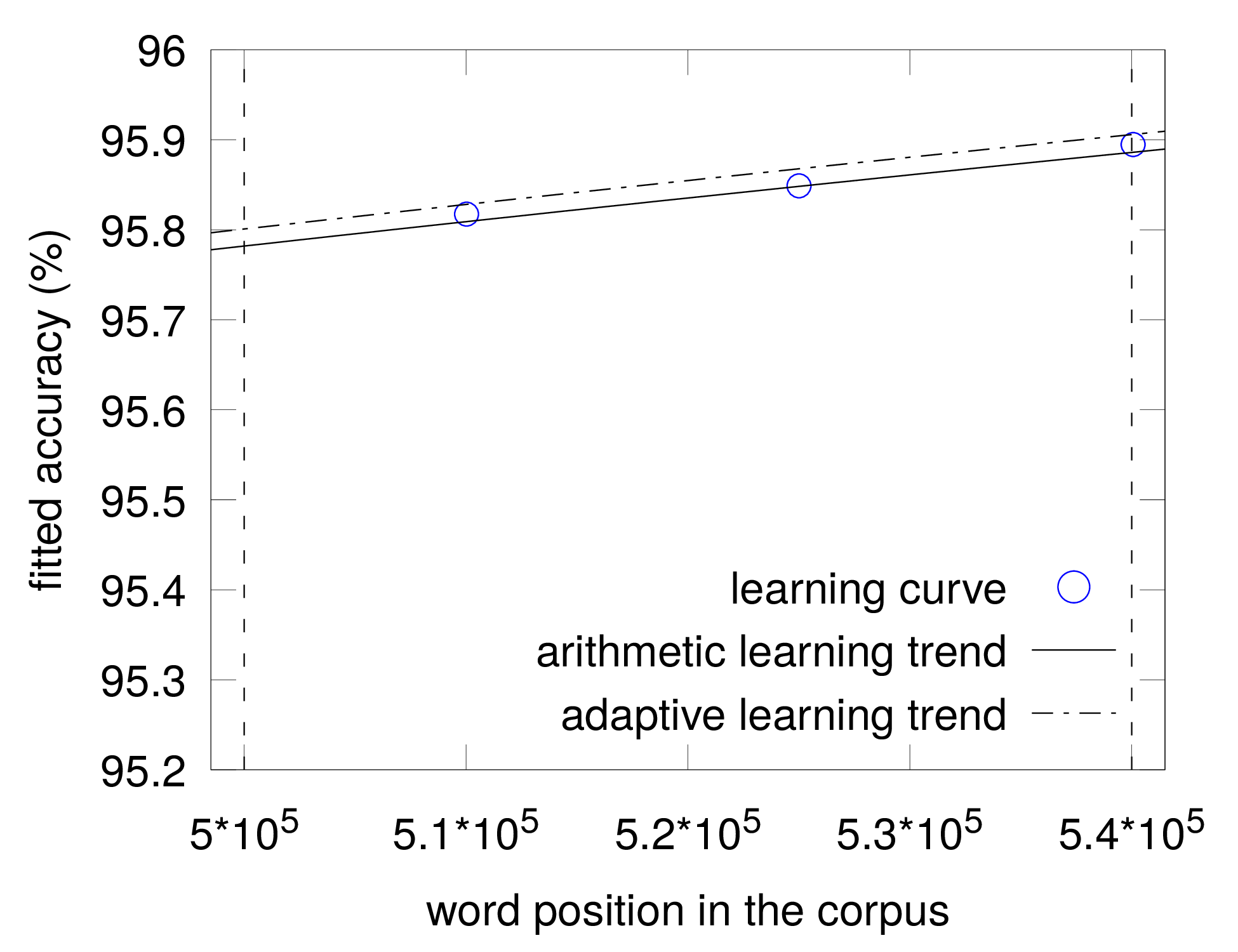}
\end{tabular}
\caption{Learning trends for fn{\sc tbl} on {\sc f}rown, using 
  uniform and adaptive step functions.}
\label{fig-trends-constant-adaptive-step-nlls-fnTBL-Frown-5000-800000-port-0.01}
\end{center}
\end{figure}

We can hence categorize the step functions according to their ability
to minimize, with regard to a {\sc port} value, the amount of training
data to be added in each iteration of a progressive sampling
process. Thus, once a {\sc port} value $\varrho$ has been determined,
the step function we are looking for is given by $\sigma[\varrho]$. We
turn again to the running example to illustrate the potential of this
outcome in
Fig.~\ref{fig-trends-constant-adaptive-step-nlls-fnTBL-Frown-5000-800000-port-0.01},
including both a general and a zoom view. The learning curve is the
same as that considered in
Fig.~\ref{fig-trace-and-level-sequence-nlls-fnTBL-Frown-5000-800000},
the observations of which we now compare with two learning trends
whose levels correspond approximately to the same position in the
corpus ($\sim4.95*10^5$), using identical kernel size ($5*10^3$) and
computed from different step functions: a uniform spacing
$\sigma=5*10^3$ and an adaptive one $\sigma[0.01]$. The latter
provides a better use of the training process by reducing the number
of cases without appreciably affecting the quality of the estimates
for accuracy. In any event, it is important to note that the
correctness has been stated from the working hypotheses, which relate
to an ideal conceptualization of the learning curves that may be
subject to variations in practice. We therefore need to analyze
mechanisms for achieving robustness of sampling.

\subsection{Robustness}

This study requires the review of our working hypotheses in order to
accommodate the notion of irregular observation in real-world {\sc
  ml}. We then assume that learning curves are positive definite and
upper bounded by 100, conditions guaranteed, but only quasi-strictly
increasing and concave. These are our \textit{testing hypotheses} and,
given that the questions to be addressed are common, we entrust the
treatment of robustness to the mechanisms intended to enhance it in
the definition of the halting condition. On this point, although any
of the solutions in the state of the art could be applied, we
raise the subject in the context of the \textit{layered convergence
  criterion}~\citep{VilaresDarribaRibadas16}, which we briefly recall
now. The choice is justified for reasons of both theoretical and
practical order, conjugating a formally correct proposal with a high
level of performance and a simple start-up. Furthermore, the approach
is easy to interpret in terms of learning traces, thus facilitating
rapid understanding. In fact, because they are not part of our
contributions but mere discussion tools, the reader can leave out the
formal definitions and results in the remainder of this Section to
focus on the intuitive interpretation accompanying them.


We identify two types of irregular observations according to their
position in relation to the \textit{working level} ({\sc wl}evel),
i.e. the iteration from which they would have a small enough impact to
work in their softening. As this depends on unpredictable factors such
as the magnitude, distribution and the very existence of these
disorders, a formal characterization is impossible and a heuristic is
necessary. Assuming that the model stabilizes as the training
advances, a way of addressing the question is categorizing the
variations induced in the monotony of the asymptotic backbone, at the
basis of the correctness for any halting condition, to locate the
level providing the first one below a given ceiling. Once the {\sc
  wl}evel passed, we are interested in estimates beyond the
\textit{prediction level} ({\sc pl}evel) marking the likely beginning
to learn trends which could feasibly predict the learning curve,
therefore not exceeding its maximum (100).

\begin{df}
\label{def-level-of-work-trace}
Let ${\mathcal A}^\pi[{\mathcal D}^{\mathcal {K}}_{\sigma}]$ be a
learning trace with asymptotic backbone $\{\alpha_i\}_{i \in
  \mathbb{N}}$, $\nu \in (0, 1)$, $\varsigma \in \mathbb{N}$ and
$\lambda \in \mathbb{N} \cup \{0\}$. We define the {\em working level}
({\sc wl}evel) {\em for} ${\mathcal A}^\pi[{\mathcal D}^{\mathcal
    {K}}_{\sigma}]$ {\em with verticality threshold} $\nu$, {\em
  slowdown} $\varsigma$ {\em and look-ahead} $\lambda$, as the
smallest $\omega(\nu,\varsigma,\lambda) \in \mathbb{N}$ verifying
\begin{equation}
\label{equation-permissible-verticality-trace}
\frac{\sqrt[\varsigma]{\nu}}{1 - \nu} \geq \frac{\abs{\alpha_{i+1} -
    \alpha_{i}}}{x_{i+1} - x_i}, \; x_i :=
\absd{{\mathcal D}_{i}}, \; \forall i \in \mathbb{N} \; \mbox{ such that } \;
\omega(\nu,\varsigma,\lambda) \leq i \leq
\omega(\nu,\varsigma,\lambda) + \lambda
\end{equation}
\noindent while the smallest $\wp(\nu,\varsigma,\lambda) \geq
\omega(\nu,\varsigma,\lambda)$ with
$\alpha_{\wp(\nu,\varsigma,\lambda)} \leq 100$ is the {\em prediction
  level} ({\sc pl}evel). Unless they are necessary for understanding,
we shall omit the parameters, referring to {\em {\sc wl}evel} by $\omega$
{\em (}resp. {\em {\sc pl}evel} by $\wp${\em )}.
\end{df}

\begin{figure}[htbp]
\begin{center}
\begin{tabular}{cc}
\hspace*{-.6cm}
\includegraphics[width=0.47\textwidth]{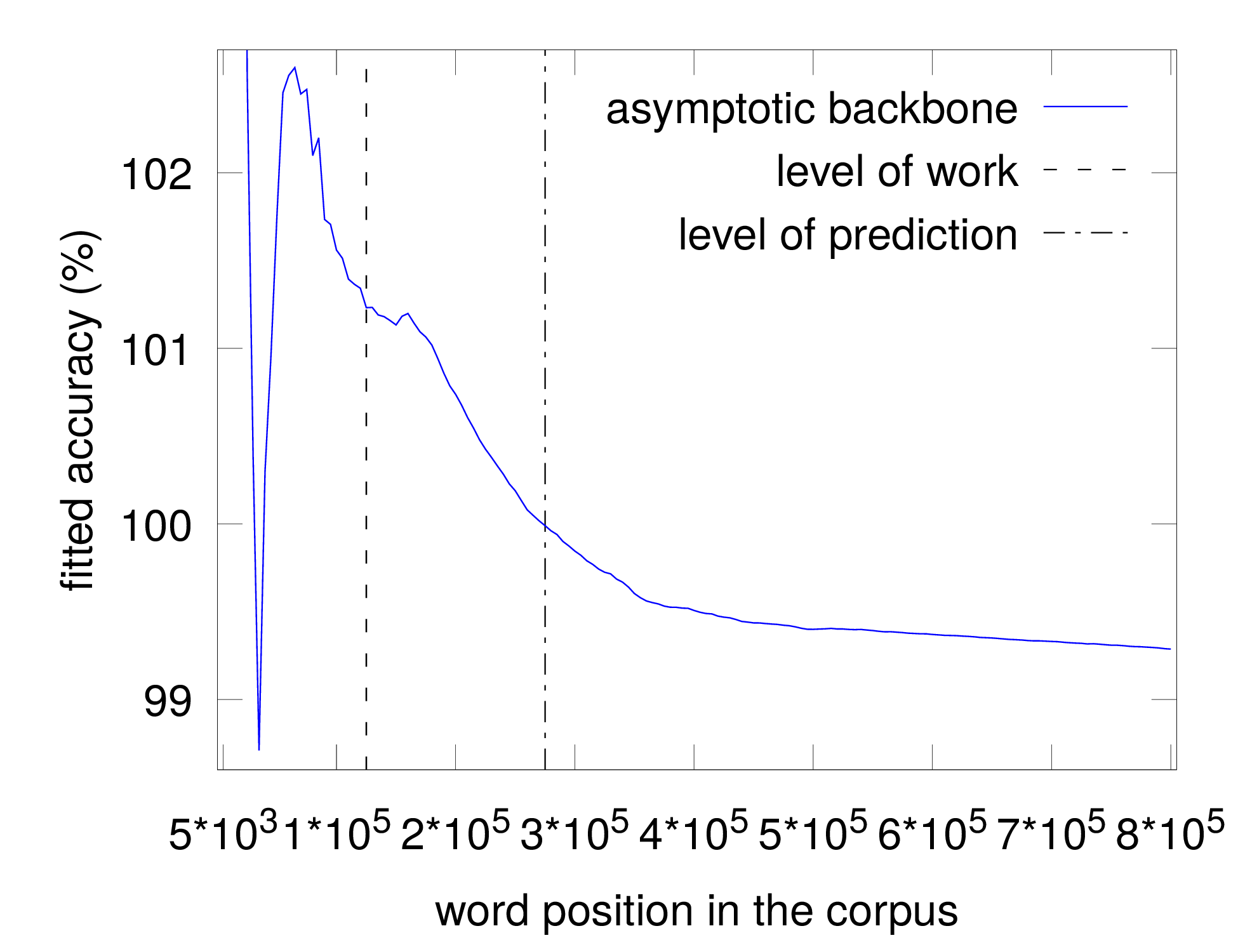} 
&
\includegraphics[width=0.47\textwidth]{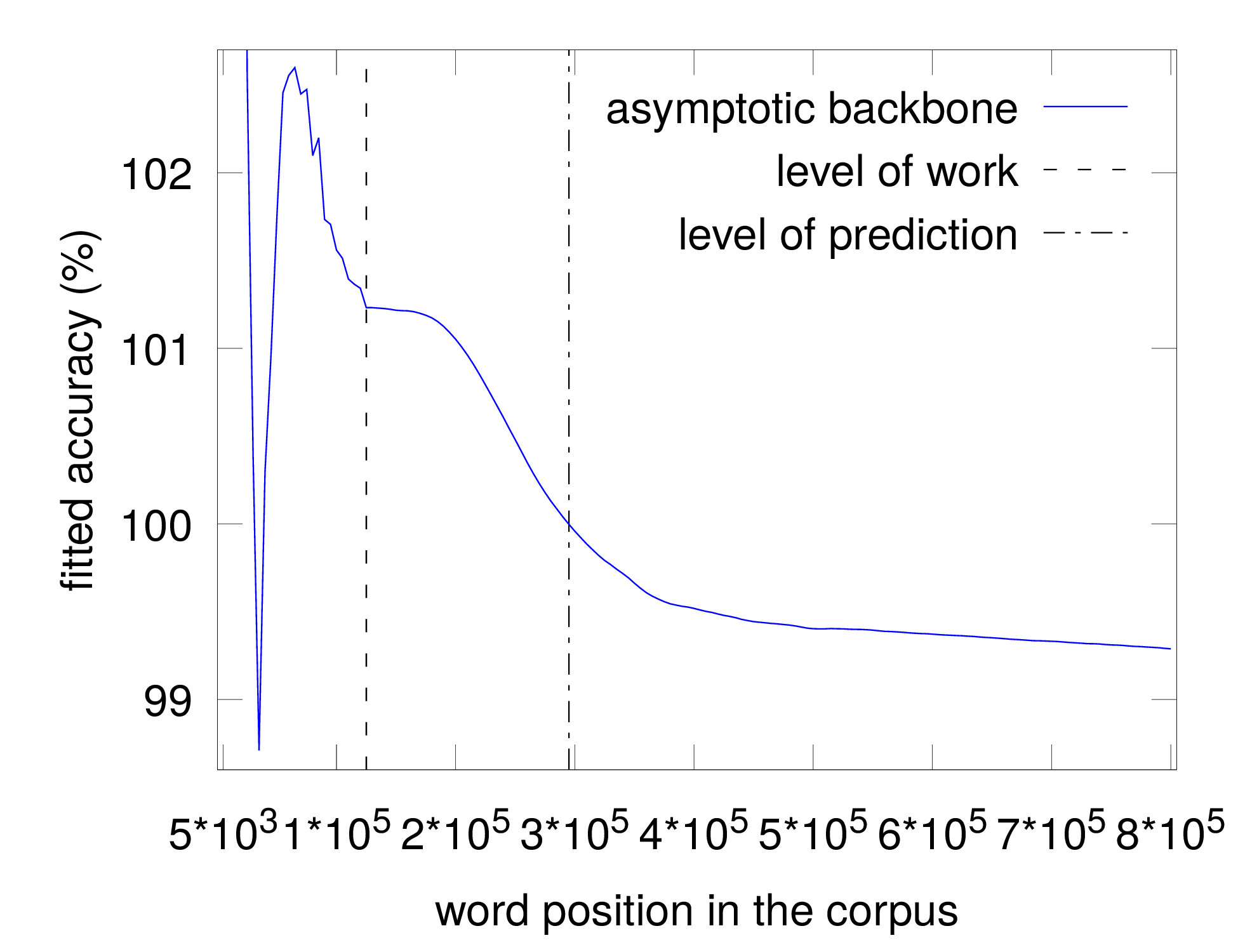}
\end{tabular}
\caption{Working and prediction levels for fn{\sc tbl} on {\sc
    f}rown.}
\label{fig-lashes-trace-and-level-sequence-nlls-fnTBL-frown-5000-800000}
\end{center}
\end{figure}

The {\sc wl}evel is the first level for which the normalized absolute
value of the slope of the line joining consecutive points on the
asymptotic backbone is less than the verticality threshold $\nu$,
which is corrected by a factor $1/\varsigma$ in order to slow down the
normalization pace, thus helping to avoid the use of infinitely small
numerals for $\nu$ in real applications. Since those tangential values
decrease together with the deviations in the monotony studied, we use
this correlation to categorize the latter, taking the look-ahead
$\lambda$ as our verification window. We then place {\sc pl}evel on
the first cycle with a learning trend below 100. In our example, the
differences of scale between disruptions in the monotony of the
asymptotic backbone before and after the {\sc wl}evel are shown in
the left-most diagram of
Fig.~\ref{fig-lashes-trace-and-level-sequence-nlls-fnTBL-frown-5000-800000} for
the parameters $\nu=2*10^{-5}$, $\varsigma=1$ and $\lambda=5$.

\subsubsection{Irregularities before the working level}

The few observations available, combined with the steep slopes of the
asymptotic backbone, have here a multiplying effect on the
fluctuations of its monotony. The use of large enough samples would
mitigate the problem, but identifying the optimal sampling size for
such purpose is equivalent to estimate the {\sc wl}evel. So, the only
effective strategy to avoid such alterations is to discard trends
associated to pre-working levels, as reflected in the left-most
diagram of
Fig.~\ref{fig-lashes-trace-and-level-sequence-nlls-fnTBL-frown-5000-800000}.

\subsubsection{Irregularities after the working level}

These should be below the verticality threshold, facilitating the
restoration of the asymptotic backbone by using an extra observation,
called \textit{anchor}, at the point of infinity of each learning
trend. Since the sum total of residuals in any of those curves is
null, this may help to neutralize irregularities. Thus, anchoring
integrates naturally into the concept of learning trace as a mechanism
to improve its robustness.

\begin{df}
\label{def-anchoring-learning-trace}
Let ${\mathcal A}^\pi[{\mathcal D}^{\mathcal {K}}_{\sigma}]$ be a
learning trace with {\em {\sc wl}evel} $\omega$, and the sequence
$\{\hat{\mathcal A}_{\ell}(\infty)\}_{\ell > \omega}$ in
$\mathbb{R}^+$. A {\em learning trend of level} $\ell > \omega$ {\em
  with anchor} $\hat{\mathcal A}_\ell(\infty)$ {\em for} ${\mathcal
  A}_{\dinfty{}}[{\mathcal D}]$ {\em using the accuracy pattern}
$\pi$, is a curve $\hat{\mathcal A}_{\ell}^\pi[{\mathcal
    D}^{\mathcal{K}}_{\sigma}] \in \pi$ fitting the observations
$\{[x_i, {\mathcal A}_{\dinfty{}}[{\mathcal D}](x_i)], \; x_i :=
\absd{\mathcal D_i} \}_{i=1}^{\ell} \cup \; [\infty, \hat{\mathcal
    A}_{\ell}(\infty)]$, whose asymptote is denoted by $y =
\hat\alpha_\ell$. When $\{\hat\alpha_\ell\}_{\ell > \omega}$ is
positive definite and converges monotonically to the asymptotic value
$\alpha_{\dinfty{}}$ of ${\mathcal A}_{\dinfty{}}[{\mathcal D}]$, we
say that $\hat{\mathcal A}^\pi[{\mathcal D}^{\mathcal {K}}_{\sigma}]
:= \{\hat{\mathcal A}_{\ell}^\pi[{\mathcal
    D}^{\mathcal{K}}_{\sigma}]\}_{\ell > \omega}$ is an {\em anchoring
  learning trace} {\em of reference} $[{\mathcal A}^\pi[{\mathcal
      D}^{\mathcal {K}}_{\sigma}], \omega]$.
\end{df}

Effectively, an anchor is treated as another observation and
located as far as the computer memory allows, which is why its use
does not modify the properties of standard learning traces. In
particular, the correctness of the proximity condition determining the
level from which the learning trends estimate the accuracy below an
error threshold~\citep{VilaresDarribaRibadas16}, is extended in a
natural way. It thus provides a simple criterion to stop a
training process, while the anchors give robustness. For ensuring its
full practical implementation, the condition is relaxed to define
it in terms of the net contribution of each learning trend to the
convergence. We then characterize the level from which such an
accuracy gain, baptized as \textit{layer of convergence}, is lower
than a ceiling fixed by the user.

\begin{thm}
\label{th-layered-correctness-trace}
{\em (Layered Correctness)} Let ${\mathcal A}^\pi[{\mathcal
    D}^{\mathcal {K}}_{\sigma}]$ be a {\em (}resp. anchoring{\em )} learning trace
with asymptotic backbone $\{\alpha_i\}_{i \in \mathbb{N}}$. We then
have that
\begin{equation}
\forall \varepsilon > 0, \; \exists n \in \mathbb{N}, \mbox{ such that
} [\chi({\mathcal A}_i^\pi[{\mathcal D}^{\mathcal {K}}_{\sigma}])
  \leq \varepsilon \Leftrightarrow i \geq n] 
\end{equation}
where $\chi({\mathcal A}_i^\pi[{\mathcal D}^{\mathcal {K}}_{\sigma}])
:= \abs{{\mathcal A}_i^\pi[{\mathcal D}^{\mathcal {K}}_{\sigma}](x_i)
  - \alpha_i}$ is the {\em layer of convergence for} ${\mathcal
  A}_i^\pi[{\mathcal D}^{\mathcal {K}}_{\sigma}], \; x_i :=
\absd{{\mathcal D}_i}, \; \forall \; i \in \mathbb{N}$.
\end{thm}

\begin{pf}
See in~\citep{VilaresDarribaRibadas16}. $\blacksquare$
\end{pf}

At this point, our sole outstanding issue is how to generate anchors
and show their effect in practice. Our thinking is based on the fact
that, having fixed a learning trend, the degree of reduction
applicable to the irregularities correlates with its residual at the
point of infinity. Extending this logic further, the closer to the
asymptote of the learning trend, the better its anchor. The problem is
that to optimize the latter we need to compute the former and vice
versa, leading us into a vicious circle. A way to avoid this is to
assign the anchor at a given level to the asymptotic value of the
previous learning trend, i.e. the last estimate available for the
accuracy resulting from a virtually infinite training process. In
return for the surrender of part of the correction potential, which
gives the strategy a conservative character, the situation is thus
unblocked to inspire the notion of \textit{canonical anchoring}.

\begin{thm}
\label{th-canonical-anchoring-trace} 
Let ${\mathcal A}^\pi[{\mathcal D}^{\mathcal {K}}_{\sigma}]$ be a
learning trace with asymptotic backbone $\{\alpha_i\}_{i \in
  \mathbb{N}}$ and $\{\hat{\mathcal A}_i(\infty)\}_{i > \omega}$ the
sequence defined from its {\em {\sc wl}evel} $\omega$ as
\begin{equation}
   \hat{\mathcal A}_{\omega+1}(\infty) :=
   \alpha_{\omega} \hspace*{.75cm}
   \hat{\mathcal A}_{i+1}(\infty) := \hat\alpha_{i} := \lim
   \limits_{x \rightarrow \infty} \hat{\mathcal A}_{i}^\pi[{\mathcal
       D}^{\mathcal{K}}_{\sigma}](x)
\end{equation}
\noindent with $\hat{\mathcal A}_{i}^\pi[{\mathcal
    D}^{\mathcal{K}}_{\sigma}]$ a curve fitting $\{[x_j, {\mathcal
    A}_{\dinfty{}}[{\mathcal D}](x_j)], \; x_j := \absd{\mathcal D_j}
\}_{j=1}^{i} \cup \; [\infty, \hat{\mathcal A}_{i}(\infty)]$, $\forall
i > \omega$. Then $\alpha_{\omega + i} \leq \hat\alpha_{\omega + i}$
{\em (resp.}  $\alpha_{\omega + i} \geq \hat\alpha_{\omega +
  i}${\em)}, $\forall i \in \mathbb{N}$, when $\{\alpha_i\}_{i \in
  \mathbb{N}}$ is decreasing {\em (resp.}  increasing{\em)}. Also,
$\{\hat{\mathcal A}_i^\pi[{\mathcal D}^{\mathcal {K}}_{\sigma}]\}_{i >
  \omega}$ is an anchoring learning trace of reference $[{\mathcal
    A}^\pi[{\mathcal D}^{\mathcal {K}}_{\sigma}],\omega]$, with
$\{\hat{\mathcal A}_i(\infty)\}_{i > \omega}$ its {\em canonical
  anchors}.
\end{thm}

\begin{pf}
See in~\citep{VilaresDarribaRibadas16}. $\blacksquare$
\end{pf}   

The effect of canonical anchoring in smoothing irregularities after
the {\sc wl}evel is illustrated, on our running example, in the
right-most diagram of
Fig.~\ref{fig-lashes-trace-and-level-sequence-nlls-fnTBL-frown-5000-800000}
versus its absence in the left-most one. It also shows how this
technique, due to its conservative nature, slows down the learning
convergence.

\section{The testing frame}
\label{section-testing-frame}

The focus is now on providing evidence of the interest in using our
proposal in a non-active adaptive sampling context, from both points
of view: training resources and learning costs. To support this, we
design a categorizing protocol for scheduling strategies from the
performance observed to converge below a given error threshold. The
access to a proximity condition for marking the end of a learning
process is then mandatory, a task we entrust to the layered
convergence criterion whose correctness was established in
Theorem~\ref{th-layered-correctness-trace}. We also need quality
metrics, which in turn require a specific monitoring architecture.

\subsection{The monitoring architecture}

After setting a {\sc ml} task for a learner on a training data base
${\mathcal D}$, the goal is to standardize the conditions under which
testing takes place, with a view to allow for its objective
assessment. Since we are talking about sampling efficiency, the true
location of the instance on which the intended effect fulfills,
hereinafter called \textit{convergence case} ({\sc cc}ase), should play
a key role in our proposal. This identifies our first objective.

Let us assume an accuracy pattern $\pi$, a kernel $\mathcal{K}$ and a
convergence threshold $\tau$. A way to approximate {\sc cc}ase is by
calculating the instance related to the \textit{convergence level}
({\sc cl}evel) \textit{of} ${\mathcal A}^\pi[{\mathcal D}^{\mathcal
    {K}}_{\eta}]$, a learning trace associated to the selected {\sc
  ml} task, with $\eta \in \mathbb{N}$ held to be fine enough. Namely,
the iteration in an arithmetic scheduling with common difference
defined by the uniform step function $\eta$, from which the error in
the estimates for accuracy is below $\tau$. In the absence of learning
malfunctions, the overvaluation of {\sc cc}ase is then less than
$\eta$. This provides the primary source of inspiration for our
monitoring strategy.

\subsubsection{The testing rounds}

Given $\eta \in \mathbb{N}$, our evaluation basis is the {\em run}, a
tuple
$\mathcal{E}_{\sigma}^\eta=[\mathcal{A}^\pi[\mathcal{D}^{\mathcal{K}}_{\sigma}],
  \wp_\eta,\tau]$ characterized by the convergence threshold $\tau$,
the {\sc pl}evel $\wp_\eta$ corresponding to
$\mathcal{A}^\pi[\mathcal{D}^{\mathcal{K}}_{\eta}]$ and a learning
trace $\mathcal{A}^\pi[\mathcal{D}^{\mathcal{K}}_{\sigma}]$ with step
function $\sigma \in \Sigma_\eta := \{\zeta \in \Sigma, \zeta(i) :=
\eta, \forall \, 2 \leq i < \wp_\eta\}$. We can then naturally extend
the notion of \textit{prediction} (resp. \textit{convergence})
\textit{level} to a run $\mathcal{E}_\sigma^{\eta}$ as the one of its
learning trace and denoting it by $\mbox{{\sc
    pl}evel}[\mathcal{E}_\sigma^{\eta}]$ (resp. $\mbox{{\sc
    cl}evel}[\mathcal{E}_\sigma^{\eta})$]. Runs can be grouped in what
we call a \textit{local testing frame of tolerance} $\eta$, a set
$\mathcal{L}[\mathcal{A}^\pi[\mathcal{D}^{\mathcal{K}}_{\Xi_\eta}],\wp_\eta,\tau]$
of these sharing $\wp_\eta$ and $\tau$, while the learning traces are
taken from
\begin{equation}
\mathcal{A}^\pi[\mathcal{D}^{\mathcal{K}}_{\Xi_\eta}] :=
\{\mathcal{A}^\pi[\mathcal{D}^{\mathcal{K}}_{\sigma}], \; \sigma \in
\Xi_\eta \subseteq \Sigma_\eta, \mbox{ such that} \; \eta \in \Xi_\eta
\}
\end{equation}
\noindent Thus, the testing round $\mathcal{E}_\eta^\eta :=
          [\mathcal{A}^\pi[\mathcal{D}^{\mathcal{K}}_{\eta}],
            \wp_\eta,\tau]$ used to approximate {\sc cc}ase with
          maximal overvaluation $\eta$, belongs to
          $\mathcal{L}[\mathcal{A}^\pi[\mathcal{D}^{\mathcal{K}}_{\Xi_\eta}],\wp_\eta,\tau]$
          and is baptized as its \textit{baseline run}. We are
          therefore talking about a package of items only
          distinguishable by their sampling schedule, taken from
          $\Xi_\eta$, while the examples calculated before the
          iteration $\wp_\eta$ are identical to those in the baseline
          $\mathcal{E}_\eta^\eta$. Accordingly, as $\wp_\eta$ is
          the {\sc pl}evel of the latter, the same applies for the
          rest of items. This provides a common starting point to
          measure, within a local testing frame, the training data
          set and the cycles needed for halting, precisely the
          parameters that we later use to quantify the converging
          effort. However, it is not enough to estimate costs to make
          sense of a performance metric: we also need to balance the
          goodness of fit results regarding {\sc cc}ase. This is
          possible thanks to the facility to visualize its interval
          of overvaluation from the baseline run, henceforth referred
          to as the \textit{interval of tolerance} $\eta$ and
          expressed by
\begin{equation}
[\imath(\mathcal{E}_\eta^{\eta},\mbox{{\sc
      cl}evel}[\mathcal{E}_\eta^{\eta}])-\eta,
  \imath(\mathcal{E}_\eta^{\eta},\mbox{{\sc
      cl}evel}[\mathcal{E}_\eta^{\eta}])]
\end{equation}
\noindent with $\imath({\mathcal E}_\sigma^\eta, \ell) := \mathcal{K}
+ \sum_{i=2}^{\ell} \sigma(\ell)$ matching a level $\ell$ in a run
$\mathcal{E}_\sigma^\eta \in
\mathcal{L}[\mathcal{A}^\pi[\mathcal{D}^{\mathcal{K}}_{\Xi_\eta}],\wp_{\eta},\tau]$
to the position of the associated instance in the training data base
${\mathcal D}$. So, with a view to adjust the degree of refinement
in estimating {\sc cc}ase, it is sufficient to shorten or lengthen
$\eta$.

\subsubsection{The testing scenarios}

We study a family ${\mathcal L} := \{{\mathcal
  L}^i[\mathcal{A}^{i,\pi}[\mathcal{D}^{\mathcal{K}^i}_{\Xi_\eta}],\wp_{\eta}^i,\tau^i]\}_{i
  \in I}$ of local testing frames, one for each combination $i \in I$
of training data base and learner. In order to simplify the
presentation, all of them share not only kernel $\mathcal{K}^i$
and accuracy pattern $\pi$, but also the tolerance $\eta$ fixed to
locate the {\sc cc}ase and the collection $\Xi_\eta$ of scheduling
schema to be compared. The robustness is entrusted to the use of
canonical anchors located at sufficient distance, in the case
$10^{200}$. To explore the response against temporary increases in
the learning curve, we force their presence in the runs studied. For the
purpose of raising expectations regarding their impact, the idea is to
generate them by increasing the accuracy observed at the last
iteration (level) before exceeding the case on which the corresponding
baseline run converges. Taking into account that it is the benchmark
instance to estimate {\sc cc}ase, itself an essential reference to
measure the sampling quality, any such inflation puts the robustness
of the scheduling strategy to the test. In this way, by applying
increases of $\iota\%$ in each run we introduce collections
$\hat{\mathcal L}[\iota], \; \iota \in (0,100]$ of local testing
frames from the original ${\mathcal L}$ family.



Formally, given a value $\iota \in (0,100]$ called \textit{inflation
    index}, we analyze a compilation of local testing frames $\hat{\mathcal
    L}[\iota] := \{\hat{\mathcal
    L}^i[\iota][\hat{\mathcal A}^{i,\pi}[\mathcal{D}^{\mathcal{K}^i}_{\Xi_\eta}],\hat\wp_{\eta}^i,\tau^i]\}_{i
    \in I}$ built from ${\mathcal L}$, in which
\begin{equation}
\hat{\mathcal
  L}^i[\iota][\hat{\mathcal A}^{i,\pi}[\mathcal{D}^{\mathcal{K}^i}_{\Xi_\eta}],\hat\wp_{\eta}^i,\tau^i]
:= \{\hat{\mathcal E}_\sigma^{\eta}\}_{\sigma \in \Xi_\eta} :=
\{[\hat{\mathcal A}^{i,\pi}[\mathcal{D}^{\mathcal{K}^i}_{\sigma}],
  \hat\wp_{\eta}^i,\tau^i]\}_{\sigma \in \Xi_\eta}, \; i \in I
\end{equation}
\noindent is such that the learning trace $\hat{\mathcal
  A}^{i,\pi}[\mathcal{D}^{\mathcal{K}}_{\sigma}]$ of each
$\hat{\mathcal E}_{\sigma}^{\eta}$ only differs from that of
${\mathcal E}_{\sigma}^{\eta} \in {\mathcal
  L}^i[\mathcal{A}^{i.\pi}[\mathcal{D}^{\mathcal{K}^i}_{\Xi_\eta}],\wp_{\eta}^i,\tau^i]$
in that it modifies the observation at the greatest level $\ell$
verifying
\begin{equation}
\imath(\mathcal{E}_\sigma^{\eta}, \ell) < \min
\{\imath(\mathcal{E}_{\sigma}^{\eta},\mbox{{\sc
    cl}evel}[\mathcal{E}_{\sigma}^{\eta}]),
\imath(\mathcal{E}_{\eta}^{\eta},\mbox{{\sc
    cl}evel}[\mathcal{E}_{\eta}^{\eta}])\}
\end{equation}
\noindent This applies by allocating the accuracy $\hat{\mathcal
  A}_{\dinfty{}}^i[\mathcal{D}](\imath(\hat{\mathcal
  E}_\sigma^{\eta},\ell))$ observed at cycle $\ell$, hereinafter
referred to as the \textit{inflated level}, to the value
\begin{equation}
\min_{}\{(1+ \frac{\iota}{100}) * {\mathcal
  A}_{\dinfty{}}^i[\mathcal{D}](\imath(\mathcal{E}_\sigma^{\eta},\ell)),
\alpha_{\mbox{{\sc cl}evel}[\mathcal{E}_\sigma^{\eta}]}\}, \; \iota
\in (0,100], \; \alpha_\ell := \lim \limits_{x \rightarrow \infty}
  {\mathcal A}_\ell^{i,\pi}[\mathcal{D}^{\mathcal{K}^i}_{\sigma}](x), \forall \ell
  \in \mathbb{N}
\end{equation}
\noindent We thus increase the accuracy at the inflated level by
$\iota\%$, as long as it does not surpass the one reached by a
hypothetically infinite training process, in order to keep these
artificial transitory inflations realistic. Moreover, to make the
comparison with runs in the starting local testing frame ${\mathcal
  L}^i[\mathcal{A}^{i,\pi}[\mathcal{D}^{\mathcal{K}^i}_{\Xi_\eta}],\wp_{\eta}^i,\tau^i]$
meaningful, it is necessary that $\hat\wp_{\eta}^i = \wp_{\eta}^i$,
for which it is sufficient that the inflated level $\ell >
\wp_{\eta}^i$. It is then said that $\hat{\mathcal
  L}^i[\iota][\hat{\mathcal
    A}^{i,\pi}[\mathcal{D}^{\mathcal{K}^i}_{\Xi_\eta}],\wp_{\eta}^i,\tau^i]$
is the \textit{inflated variant at a rate of $\iota\%$ for} ${\mathcal
  L}^i[\mathcal{A}^{i,\pi}[\mathcal{D}^{\mathcal{K}^i}_{\Xi_\eta}],\wp_{\eta}^i,\tau^i]$,
the type of local testing frame from which the collections
$\hat{\mathcal L}[\iota]$ are effectively built. This enlarges our
testing canvas in a relevant manner, allowing appraisal of the effect
of mismatches in the working hypotheses without compromising the rest
of the surrounding conditions.

\subsection{The learning performance metrics}

Sharing the {\sc pl}evel and a halting condition based on predictive
accuracy allows the runs in a local testing frame ${\mathcal
  L}[\mathcal{A}^\pi[\mathcal{D}^{\mathcal{K}}_{\Xi_\eta}],\wp_{\eta},\tau]$
or in any of its inflated variants $\hat{\mathcal
  L}[\iota][\hat{\mathcal
    A}^\pi[\mathcal{D}^{\mathcal{K}}_{\Xi_\eta}],\wp_{\eta},\tau]$ to
support a reliable methodology to compare their learning
performance. The first feature provides a common starting point for
the evaluation process, while the second determines its conclusion and
associates it to a {\sc cl}evel. It is therefore possible to identify
the effective testing areas together with the iterations involved,
from the same departure position and without using heuristics. So,
after having set a run, a quality metric simply needs to focus on the
cycles between a {\sc pl}evel of value $\wp_\eta$ shared in the local
testing frame and its own {\sc cl}evel.

This way, data acquisition costs are proportional to the fraction of
training data explored while, in the absence of incremental {\sc ml}
mechanisms\footnote{A model is updated from new examples and a limited
  set of previous ones. The idea is that adding or removing small
  amounts of data, it may not change much and the incrementality
  should reduce the learning effort. However, practice proves that its
  applicability is doubtful: the use scenarios may significantly vary
  according to applications~\citep{Castro18}, it is not guaranteed to
  be faster than re-training from scratch~\citep{Tsai14} and
  catastrophic forgetting~\citep{French99} can arise~\citep{Losing18}
  when the model at stake is connectionist.}, model induction ones
depend on the iterations made to that end and on the new cases added
in each cycle. Since the use of a common proximity condition
guarantees that misclassification rates are the same within a local
testing frame, error charges are irrelevant in comparative terms. We
can therefore assume, following Weiss and Tian~\citep{WeissTian08} and
for testing purposes, that overall learning costs depend only on the
data acquisition and induction ones. Moreover, as the intention is to
study those charges through different local testing frames, we are
more interested in calculating ratios than providing absolute
values. To this effect, we normalize them taking into account that
$\imath({\mathcal E}_\eta^\eta, \wp_\eta)$ is the first instance on
which a run can converge, thus quantifying a benchmark for both
settings.

Having discussed the quantitative side of our performance metrics, we
need now to integrate the qualitative one. In this sense, and always
in the context of a local testing frame and its inflated variants, we
positively evaluate any approximation for {\sc cc}ase, provided it is
located from the start of the interval of tolerance. Interpreted as an
indication that training is less efficient, the increasing distance
from the latter is penalized to an extent proportional to the costs,
and therefore also to the training resources, mentioned above. On the
other hand, a premature diagnosis is not acceptable because it
necessarily entails an error in terms of accuracy prediction. Given
these premises, we formally introduce the performance metrics.

\begin{df}
\label{def-overall-learning-cost-and-used-training-resources-ratios}
Let
$\mathcal{E}_\sigma^\eta \in
\mathcal{L}[\mathcal{A}^\pi[\mathcal{D}^{\mathcal{K}}_{\Xi_\eta}],\wp_{\eta},\tau]$
be a run in a local testing frame with baseline
${\mathcal E}_\eta^\eta$, and ${\mathcal H}$ an halting condition. We
define the {\em data acquisition} (resp. {\em induction}) {\em
  cost saving ratio of} $\mathcal{E}_\sigma^\eta$ {\em for} ${\mathcal H}$
as
\begin{equation}
\label{eq-data-acquisition-cost-saving-ratio}
\mbox{\sc dacsr}({\mathcal E}_\sigma^\eta,{\mathcal H}) :=
\left\{ \begin{array}{ll} \frac{\imath({\mathcal E}_\eta^\eta,
    \wp_\eta)}{\imath({\mathcal E}_\sigma^\eta, \mbox{\em {\sc
        cl}evel}[{\mathcal E}_\sigma^\eta])} & \mbox{if } \;
  \delta({\mathcal E}_\sigma^\eta, \mathcal{H}) \in [-\eta, \infty) \\ 0
    & \mbox{otherwise}
\end{array} \right.
\end{equation}

\begin{equation}
\label{eq-induction-cost-saving-ratio}
(\mbox{resp. } \mbox{\sc icsr}({\mathcal E}_\sigma^\eta,{\mathcal H}) :=
\frac{\sum_{\ell=1}^{\wp_\eta}\imath({\mathcal E}_\eta^\eta,\ell)}{\sum_{\ell=1}^{\wp_\eta - 1}\imath({\mathcal E}_\eta^\eta,\ell) + \sum_{\ell=\wp_\eta}^{\mbox{\scriptsize\em {\sc cl}evel}[{\mathcal E}_\sigma^\eta]}\imath({\mathcal E}_\sigma^\eta,\ell)} )
\end{equation}
\noindent with
$\delta({\mathcal E}_\sigma^\eta, \mathcal{H}) := \imath({\mathcal
  E}_\sigma^\eta, \mbox{\em {\sc cl}evel}[{\mathcal E}_\sigma^\eta]) -
\imath({\mathcal E}_\eta^\eta, \mbox{\em {\sc cl}evel}[{\mathcal
  E}_\eta^\eta])$ the {\em discrepancy distance of}
${\mathcal E}_\sigma^\eta$ {\em for} ${\mathcal H}$. From which, we
introduce the {\em overall learning cost saving ratio} of
$\mathcal{E}_\sigma^\eta$ {\em for} ${\mathcal H}$ as

\begin{equation}
\label{eq-overall-learning-cost-saving-ratio}
\mbox{\sc lcsr}({\mathcal E}_\sigma^\eta,{\mathcal H}) :=
\mbox{\sc dacsr}({\mathcal E}_\sigma^\eta,{\mathcal H}) \ast
\mbox{\sc icsr}({\mathcal E}_\sigma^\eta,{\mathcal H})
\end{equation}  

\end{df}

In the context of a local testing frame, these metrics are null when
the run ${\mathcal E}_\sigma^\eta$ converges before the interval
$[\imath(\mathcal{E}_\eta^{\eta},\mbox{{\sc
    cl}evel}[\mathcal{E}_\eta^{\eta}])-\eta,
\imath(\mathcal{E}_\eta^{\eta},\mbox{{\sc
    cl}evel}[\mathcal{E}_\eta^{\eta}])]$ of tolerance $\eta$ for
locating {\sc cc}ase. Otherwise, the greater the discrepancy distance
the lower their value, reaching a maximum of 1 if
$\mbox{{\sc cl}evel}[{\mathcal E}_\sigma^\eta] = \wp_\eta$, i.e. if
$\mbox{{\sc cl}evel}[{\mathcal E}_\sigma^\eta] = \mbox{{\sc
    pl}evel}[{\mathcal E}_\sigma^\eta] = \mbox{{\sc pl}evel}[{\mathcal
  E}_\eta^\eta]$. Namely, when the learning process halts at the same
time as the predictions are judged reliable, thereby signalling the
least costly convergence process within such an interval of
tolerance. That way, we have not only a realistic saving quota for the
overall learning cost ({\sc lcsr}) but also for the training resources
used ({\sc dacsr}), precisely the two magnitudes on which we focus our
attention.

\section{The experiments}
\label{section-experiments}

As mentioned earlier, the focus here is on learners associated to {\sc
  ml}-based tagger generation, a demanding task in the domain of {\sc
  nlp}. It is thus necessary to introduce the linguistic resources and
the testing space.


\subsection{The linguistics resources}

Corpora and {\sc pos} tagger generators are selected from the most
popular ones, as training data and learners respectively, the former
together with their tag-sets are:

\begin{enumerate}
\item The Wall Street Journal section in the {\sc p}enn {\sc
  t}reebank~\citep{Marcus1999}, with over 1,170,000 words.


\item The Freiburg-Brown ({\sc f}rown) of American
  English~\citep{Hinrichs2010}, with over 1,165,000 words.
\end{enumerate}

\noindent where {\sc p}enn is annotated with {\sc pos} tags as well as
syntactic structures. By stripping it of the latter, it can be used to
train {\sc pos} tagging systems. In order to ensure well-balanced
corpora, we also have scrambled them at sentence level before testing.

In the case of taggers, we focus on systems built from supervised
learning, which make it possible to work with predefined tag-sets,
thereby facilitating both the evaluation and the comprehension of the
results in contrast with unsupervised techniques:

\begin{enumerate}

\item In the category of stochastic methods and as a representative of
  the \textit{hidden M\'arkov models} ({\sc hmm}{\footnotesize s}), we
  chose {\sc t}n{\sc t}~\citep{Brants2000}. We also include the {\sc
    t}ree{\sc t}agger~\citep{Schmid1994}, which uses decision trees to
  generate the {\sc hmm}, and {\sc m}orfette~\citep{Chrupala2008}, an
  averaged perceptron approach~\citep{Collins2002}. To illustrate the
  \textit{maximum entropy models} ({\sc mem}{\footnotesize s}), we
  work with {\sc mxpost}~\citep{Ratnaparki1996} and {\sc o}pen{\sc
    nlp} {\sc m}ax{\sc e}nt~\citep{Toutanova2003}. Finally, the {\sc
    s}tanford {\sc pos} tagger~\citep{Toutanova2003} combines features
  of {\sc hmm}{\footnotesize s} and {\sc mem}{\footnotesize s} using a
  \textit{conditional M\'arkov model}.

\item Under the heading of other methods we take fn{\sc
  tbl}~\citep{Ngai2001}, an update of {\sc b}rill~\citep{Brill1995a},
  as an example of transformation-based learning. For memory-based
  strategies, the chosen representative is the \textit{memory-based tagger}
  ({\sc mbt})~\citep{Daelemans1996a}, while {\sc
    svmt}ool~\citep{Gimenez2004} illustrates behaviour with
  respect to support vector machines. Finally, we use a
  perceptron-based training method with look-ahead, through {\sc
    lapos}~\citep{Tsuruoka2011}.
\end{enumerate}
\noindent This selection ensures good coverage of the linguistic
resources with a view to test our proposal, thus providing a solid and
representative basis for our experiments.

\subsection{The testing space}

Under our guidelines, we consider a local testing frame together with
its inflated variant at a rate of $1\%$ for each combination $i \in I$
of corpus and tagger, grouping them in the collections
\[{\mathcal L}
:= \{{\mathcal L}^i[\mathcal{A}^\pi[\mathcal{D}^{{\mathcal
        K}^i}_{\Xi_\eta}],\wp_{\eta}^i,\tau^i]\}_{i \in I} \; \mbox{ and } \; 
\hat{\mathcal L}[1] := \{\hat{\mathcal L}^i[1][\hat{\mathcal
    A}^{i,\pi}[\mathcal{D}^{{\mathcal
        K}^i}_{\Xi_\eta}],\wp_{\eta}^i,\tau^i]\}_{i \in I}
\]
\noindent where the tolerance $\eta$ and the size of the kernels
${\mathcal K}^i$ are both fixed to $5*10^3$, while a power law family
parameterized by the trust region method~\citep{Branch1999} is chosen
as accuracy pattern $\pi$. The set $\Xi_\eta$ of adaptive sampling
schedules to be compared includes our own ({\sc colts}), as well as
the geometric and the arithmetic ones, the latter indicating the
baseline runs. The values for $\{\wp^i_\eta\}_{i \in I}$ (resp.
$\{\tau^i\}_{i \in I}$), i.e. the {\sc pl}evels of the baselines
(resp. the convergence thresholds), are calculated from the parameters
$\nu=2*10^{-5}$, $\varsigma=1$ and $\lambda=5$ (resp. the same) used
by Vilares \textit{et al.}~\cite{VilaresDarribaRibadas16}.  Hence each
local testing frame and inflated variant is composed of three runs,
all of them operating under identical testing conditions. It is
important to note that geometric scheduling generally results in a
reduction of induction costs, but at the price of increasing data
acquisition ones and relaxing the precision. Meanwhile, the arithmetic
approach shows an opposite behavior, allowing optimum setting for the
training resources used, which justifies our decision to choose it for
defining the baseline runs. So, these two strategies represent the
foreseeable extremes as regards performance in practical adaptive
sampling, thus justifying their inclusion in our experimental frame.

Having defined the main testing structure, we address three aspects
supporting the significance of the trials in our case study. The first
relates to the appropriate exploitation of the training
resources. Thus, as phrases are the smallest grammatical units with
concrete sense, samples should be aligned to the sentential
distribution of the text. The second concerns the practical utility of
the generated models, which depends on both the {\sc lcsr} metric
being well-defined within the scope of the corpora and the reduction
of variability phenomena. Finally, we tackle the model optimization,
i.e. the refinement of schedule setting in each run.

\subsubsection{Sampling fitting to sentence level}

In order to guarantee the relevance of our experimental results, the
best training conditions for {\sc pos} tagging should be
provided. This includes avoiding dysfunctions resulting from sentence
truncation, which requires the use of a particular class of learning
scheme specifically adapted to sentence level. So, given a corpus
$\mathcal D$ with kernel $\mathcal K$ and a step function $\sigma$, we
build the individuals $\{\mathcal D_i\}_{i \in \mathbb{N}}$ with
${\mathcal D}_i := \sentences{{\mathcal W}_i}$ such that
\begin{equation}
\begin{array}{l}
{\mathcal W}_1 := {\mathcal K} \mbox{ and } {\mathcal W}_i :=
{\mathcal W}_{i-1} \cup {\mathcal I}_{i}, \; \mathcal I_i \subset
{\mathcal D} \setminus {\mathcal W}_{i-1}, \; \absd{{\mathcal I}_{i}}
:= \sigma(i), \; \forall i \geq 2
\end{array}
\end{equation}
\noindent where $\sentences{{\mathcal W}_{i}}$ denotes the minimal set
of sentences including the set of words ${\mathcal W}_{i}$. Such a fit
has no impact on the foundations of the proposal and allows us to reap
the maximum benefit from the training process.

\subsubsection{Scope and stability of sampling}

Since the corpora considered are finite, it only makes sense to study
a local testing frame ${\mathcal
  L}[\mathcal{A}^\pi[\mathcal{D}^{\mathcal{K}}_{\Xi_\eta}],\wp_{\eta},\tau]$
when {\sc cc}ase, i.e. the instance on which the testing condition
verifies, is within their boundaries. That is to say, when the
alignment below the threshold $\tau$ occurs in that context. With the
aim of adapting to this practical constraint, we limit the scope in
measuring the layer of convergence as introduced in
Theorem~\ref{th-layered-correctness-trace} for a learning trend
${\mathcal A}_i^\pi[{\mathcal D}^{\mathcal {K}}_{\sigma}]$. Formally,
the asymptotic value $\alpha_i$ is replaced by the one reached at the
last case for which an observation is available in the corpus we are
working on. So, if $\sentencesw{\ell}$ denotes the position of the
first sentence-ending beyond the $\ell$-th word, the layer of
convergence for ${\mathcal A}_i^\pi[{\mathcal D}^{\mathcal
    {K}}_{\sigma}]$ is now expressed by
\begin{equation}
\chi^{\ell}({\mathcal A}_i^\pi[{\mathcal D}^{\mathcal {K}}_{\sigma}]) :=
\abs{{\mathcal A}_i^\pi[{\mathcal D}^{\mathcal
      {K}}_{\sigma}](\absd{{\mathcal D}_i}) - \pmb{{\mathcal
    A}_i^\pi[{\mathcal D}^{\mathcal
      {K}}_{\sigma}](\sentencesw{\ell})}}
\end{equation}
\noindent with $\ell=8*10^5$ and the updated term represented in bold
font. In order to confer stability on our measures, a $k$-fold cross
validation~\citep{Clark2010} is applied with $k$=10, a commonly used
value in {\sc pos} tagging
evaluation~\citep{Daelemans1996a,Giesbrecht2009}.

\subsubsection{Parameter tuning}

The performance of a sampling schedule relies heavily on the learning
process being studied, in such a way that it may even challenge our
initial expectations. So, the uniform strategy generally outperforms
the geometric one in terms of training resources used because the
efficiency in approximating the real learning curve is higher. By
contrast, since the number of cycles required to converge can be very
large, it often entails significant model induction costs unless an
incremental learning facility prevents overlapping of training data in
successive iterations, which is a technology far from being
operational. The same goes for error costs when the issue is the
learning utility~\citep{WeissTian08}.

Having fixed a local testing frame
${\mathcal
  L}[\mathcal{A}^\pi[\mathcal{D}^{\mathcal{K}}_{\Xi_\eta}],\wp_{\eta},\tau]$,
we therefore need a protocol for tuning the step function
$\sigma \in \Xi_\eta$ in each run ${\mathcal E}_{\sigma}^{\eta}$ when
$\sigma \neq \eta$. With the aim of offering meaningful results, the
starting conditions should be as close as possible to those of the
baseline run ${\mathcal E}_{\eta}^{\eta}$, which associates a quasi-optimal
{\sc cl}evel. Since
$\sigma \in \Xi_\eta \subseteq \Sigma_\eta := \{\zeta \in \Sigma,
\zeta(i) := \eta, \forall \; 2 \leq i < \wp_\eta\}$, this goes on to
choose $\sigma$ in such a way that $\sigma(\wp_\eta) = \eta$. The
settings thus calculated are also applied on the inflated variant
$\hat{\mathcal L}[1][\hat{\mathcal
  A}^\pi[\mathcal{D}^{\mathcal{K}}_{\Xi_\eta}],\wp_{\eta},\tau]$.

\paragraph{Setting the geometric scheduling}

The step function $\sigma$ is defined from a common ratio $\rho
\in \mathbb{R}^+ \setminus (0, 1]$, which is why we denote it as
  $\sigma := \sigma[\rho]$, and the condition to be verified is
  expressed by
\begin{equation}
\eta = \sigma[\rho](\wp_\eta) := \imath({\mathcal E}_{\sigma[\rho]}^\eta,
\wp_\eta + 1) - \imath({\mathcal E}_{\sigma[\rho]}^\eta, \wp_\eta) =
\imath({\mathcal E}_{\sigma[\rho]}^\eta, \wp_\eta) \ast (\rho - 1) :=
\imath({\mathcal E}_\eta^\eta, \wp_\eta) \ast ( \rho - 1)
\end{equation}
\noindent while the common ratio we are looking for is
\begin{equation}
\rho := \frac{\eta + \imath({\mathcal E}_\eta^\eta, \wp_\eta)}
             {\imath({\mathcal E}_\eta^\eta, \wp_\eta)}
\end{equation}

\paragraph{Setting the adaptive scheduling}

The step function $\sigma$ is defined from a {\sc port} parameter
$\varrho \in (0, 1]$, so that we denote it by $\sigma :=
  \sigma[\varrho]$. The main problem in this respect is that we are
  talking about non-fixed sequencing schedulings, for which the step
  at each level depends not only on the set of observations available
  at that moment but also on the accuracy pattern $\pi$ used for
  approximating the learning trends. Namely, unlike geometric
  scheduling, it is not possible to statically solve for $\varrho$ the
  equation $\sigma[\varrho](\wp_\eta) = \eta$. An approach that is
  also adaptive, in accordance with the nature of this sampling
  strategy, is therefore necessary. So, having taken an initial
  tentative {\sc port} value $\psi$, we adjust it to a new one
  $\varrho[\psi]$ in such a way that $\sigma[\varrho[\psi]](\wp_\eta)=
  \eta$. In principle, it would be sufficient for this to apply the
  transformation
\begin{equation}
\label{eq-first-PORT-approximation-adaptive}
\varrho[\psi] := \frac{\eta}{\sigma[\psi](\wp_\eta)}, \mbox{ with } \psi \in (0,1] 
\end{equation}  
\noindent but we must take into account that $\sigma[\psi](\wp_\eta)$
should not exceed the size of the remaining training data base
${\mathcal D}$ from the level $\wp_\eta$ and that the {\sc port} is
defined in $(0, 1]$. The definitive transformation to be used is then
\begin{equation}
\label{eq-final-PORT-approximation-adaptive}
\varrho[\psi] := \min \{\frac{\eta} {\min \{\sigma[\psi](\wp_\eta),
  \absd{{\mathcal D}-{\mathcal D}_{\wp_\eta}}\}}, 1\}, \mbox{ with }
\psi \in (0,1]
\end{equation}
\noindent We choose the intermediate value $\psi = 0.5$ to start the
approximation process. This allows us to obtain an initial idea of the
proposal's potential, leaving the study of the impact on performance
due to {\sc port} variations for later.


\subsection{Analysis of the results}

The detail of the monitoring can be found separately for each local
testing frame in ${\mathcal L}$ and $\hat{\mathcal L}[1]$, on
Tables~\ref{table-runs-without-inflation}
and~\ref{table-runs-with-inflation} respectively. That includes the
{\sc pl}evel, common to all its runs and expressed by both the numeric
value (\#) and the position ($\imath$) of the related instance (word)
in the corpus, and also the results for the quality metrics {\sc lcsr}
and {\sc dacsr} on each one of those runs. The first indicator
signals, as already pointed out, and once an error convergence
threshold has been fixed, the iteration from which performance
prediction on the learner is presumed realistic. Meanwhile, {\sc lcsr}
and {\sc dacsr} give an objective view of the saving effort with
respect to overall learning cost and training resources used,
respectively. We thus hope that {\sc colts}-based runs equal
the results of the baselines for {\sc dacsr}, while reaching the best
ones for {\sc lcsr}.

\begin{table*}
\begin{center}
\begin{footnotesize}
\begin{tabular}{@{\hspace{0pt}}l@{\hspace{3pt}}l@{\hspace{2pt}}l@{\hspace{2pt}}
                c@{\hspace{0pt}}c@{\hspace{3pt}}  
                r@{\hspace{0pt}}c@{\hspace{3pt}}  
                c@{\hspace{0pt}}c@{\hspace{3pt}}  
                c@{\hspace{0pt}}c@{\hspace{2pt}}  
                c@{\hspace{0pt}}c@{\hspace{3pt}}  
                c@{\hspace{0pt}}c@{\hspace{2pt}}  
                c@{\hspace{0pt}}c@{\hspace{2pt}}  
                c@{\hspace{0pt}}c@{\hspace{3pt}}  
                c@{\hspace{0pt}}c@{\hspace{2pt}}  
                c@{\hspace{0pt}}c@{\hspace{2pt}}  
                c@{\hspace{0pt}}c@{\hspace{2pt}}  
                c@{\hspace{0pt}}}                 

\hline
& & & \multicolumn{3}{c}{\bf{\scshape pl}evel} & & \boldmath$\mathbf\tau$ & & \multicolumn{5}{c}{\bf Baseline} & & \multicolumn{5}{c}{\bf Geometric} & & \multicolumn{5}{c}{\bf{\scshape colts}} \rule{0pt}{2.5ex} \\

\cline{1-2} \cline{4-6} \cline{8-8} \cline{10-14} \cline{16-20} \cline{22-26}

& & & {\footnotesize \bf \#} & & {\footnotesize $\mathbf\imath$\hspace*{12pt}} & & & & \boldmath$\mathbf\eta$ & & \bf{\scshape dacsr} & & \bf{\scshape lcsr} & & \boldmath$\rho$ & & \bf{\scshape dacsr} & & \bf{\scshape lcsr} & & \boldmath$\varrho[0.5]$ & & \bf{\scshape dacsr} & & \bf{\scshape lcsr}\hspace*{-3pt} \rule{0pt}{2.5ex} \\

\cline{1-2} \cline{4-6} \cline{8-8} \cline{10-10} \cline{12-12} \cline{14-14} \cline{16-16} \cline{18-18} \cline{20-20} \cline{22-22} \cline{24-24} \cline{26-26}

\multirow{7}{*}{\begin{sideways}{\bf \textsc{f}rown}\end{sideways}} & \textsc{lapos} & & 18 & & 90,000 & & 1.27 & &   \em 5,000 & &  \em 0.3913 & &  \em 0.0619 & &  1.056 & &  0.3765 & &  0.0876 & & \bf  0.048 & & \bf  0.3487 & & \bf  0.1175\rule{0pt}{2.75ex} \\
& \textsc{m}ax\textsc{e}nt & & 32 & & 160,004 & & 1.70 & &   \em 5,000 & &  \em 0.6400 & &  \em 0.2651 & &  1.031 & &  0.6304 & &  0.2901 & & \bf  0.023 & & \bf  0.6171 & & \bf  0.3757\rule{0pt}{2.25ex} \\
& \textsc{m}orfette & & 20 & & 100,009 & & 1.43 & &   \em 5,000 & &  \em 0.4166 & &  \em 0.0744 & &  1.050 & &  0.4162 & &  0.1093 & & \bf  0.039 & & \bf  0.3851 & & \bf  0.1442\rule{0pt}{2.25ex} \\
& \textsc{mxpost} & & 22 & & 110,017 & & 2.84 & &   \em 5,000 & &  \em 0.7334 & &  \em 0.3991 & &  1.045 & &  0.7012 & &  0.3783 & & \bf  0.039 & & \bf  0.7063 & & \bf  0.4925\rule{0pt}{2.25ex} \\
& \textsc{s}tanford & & 29 & & 145,014 & & 1.91 & &   \em 5,000 & &  \em 0.7631 & &  \em 0.4480 & &  1.034 & &  0.7625 & &  0.4698 & & \bf  0.024 & & \bf  0.7183 & & \bf  0.5100\rule{0pt}{2.25ex} \\
& \textsc{svmt}ool & & 46 & & 230,005 & & 1.41 & &   \em 5,000 & &  \em 0.8679 & &  \em 0.6557 & &  1.022 & &  0.8788 & &  0.6890 & & \bf  0.016 & & \bf  0.8449 & & \bf  0.7103\rule{0pt}{2.25ex} \\
& \textsc{t}n\textsc{t} & & 19 & & 95,018 & & 1.51 & &   \em 5,000 & &  \em 0.4419 & &  \em 0.0888 & &  1.053 & &  0.4188 & &  0.1108 & & \bf  0.038 & & \bf  0.4080 & & \bf  0.1611\rule{0pt}{2.25ex} \\
\cline{1-2} \cline{4-6} \cline{8-8} \cline{10-10} \cline{12-12} \cline{14-14} \cline{16-16} \cline{18-18} \cline{20-20} \cline{22-22} \cline{24-24} \cline{26-26}

\multirow{10}{*}{\begin{sideways}{\bf \textsc{p}enn}\end{sideways}} & fn\textsc{tbl} & & 19 & & 95,007 & & 0.58 & &   \em 5,000 & &  \em 0.3334 & &  \em 0.0383 & &  1.053 & &  0.3226 & &  0.0622 & & \bf  0.058 & & \bf  0.3265 & & \bf  0.1045\rule{0pt}{2.75ex} \\
& \textsc{lapos} & & 13 & & 65,003 & & 0.93 & &   \em 5,000 & &  \em 0.4814 & &  \em 0.1159 & &  1.077 & &  0.4765 & &  0.1491 & & \bf  0.086 & & \bf  0.4879 & & \bf  0.2301\rule{0pt}{2.25ex} \\
& \textsc{m}ax\textsc{e}nt & & 19 & & 95,007 & & 0.60 & &   \em 5,000 & &  \em 0.3585 & &  \em 0.0476 & &  1.053 & &  0.3575 & &  0.0780 & & \bf  0.062 & & \bf  0.3626 & & \bf  0.1292\rule{0pt}{2.25ex} \\
& \textsc{mbt} & & 15 & & 75,035 & & 1.66 & &   \em 5,000 & &  \em 0.4287 & &  \em 0.0817 & &  1.066 & &  0.4344 & &  0.1200 & & \bf  0.054 & & \bf  0.4164 & & \bf  0.1673\rule{0pt}{2.25ex} \\
& \textsc{m}orfette & & 15 & & 75,035 & & 0.52 & &   \em 5,000 & &  \em 0.3573 & &  \em 0.0475 & &  1.066 & &  0.3583 & &  0.0778 & & \bf  0.094 & & \bf  0.3598 & & \bf  0.1265\rule{0pt}{2.25ex} \\
& \textsc{mxpost} & & 17 & & 85,013 & & 1.40 & &   \em 5,000 & &  \em 0.5862 & &  \em 0.2062 & &  1.059 & &  0.5635 & &  0.2215 & & \bf  0.061 & & \bf  0.5739 & & \bf  0.3215\rule{0pt}{2.25ex} \\
& \textsc{s}tanford & & 18 & & 90,031 & & 0.98 & &   \em 5,000 & &  \em 0.6001 & &  \em 0.2207 & &  1.055 & &  0.5841 & &  0.2402 & & \bf  0.050 & & \bf  0.5311 & & \bf  0.2754\rule{0pt}{2.25ex} \\
& \textsc{svmt}ool & & 26 & & 130,008 & & 1.25 & &   \em 5,000 & &  \em 0.7428 & &  \em 0.4139 & &  1.039 & &  0.7389 & &  0.4334 & & \bf  0.028 & & \bf  0.6909 & & \bf  0.4716\rule{0pt}{2.25ex} \\
& \textsc{t}n\textsc{t} & & 12 & & 60,015 & & 0.51 & &   \em 5,000 & &  \em 0.2609 & &  \em 0.0188 & &  1.083 & &  0.2574 & &  0.0379 & & \bf  0.087 & & \bf  0.2508 & & \bf  0.0593\rule{0pt}{2.25ex} \\
& \textsc{t}ree\textsc{t}agger & & 12 & & 60,015 & & 1.32 & &   \em 5,000 & &  \em 0.2728 & &  \em 0.0215 & &  1.083 & &  0.2574 & &  0.0379 & & \bf  0.066 & & \bf  0.2500 & & \bf  0.0590\rule{0pt}{2.25ex} \\
\cline{1-2} \cline{4-6} \cline{8-8} \cline{10-10} \cline{12-12} \cline{14-14} \cline{16-16} \cline{18-18} \cline{20-20} \cline{22-22} \cline{24-24} \cline{26-26}
\hline
\end{tabular}
\end{footnotesize}
\end{center}
\caption{Monitoring of local testing frames without inflation}
\label{table-runs-without-inflation}
\end{table*}

\begin{table*}
\begin{center}
\begin{footnotesize}
\begin{tabular}{@{\hspace{0pt}}l@{\hspace{3pt}}l@{\hspace{2pt}}l@{\hspace{2pt}}
                c@{\hspace{0pt}}c@{\hspace{3pt}}  
                r@{\hspace{0pt}}c@{\hspace{3pt}}  
                c@{\hspace{0pt}}c@{\hspace{3pt}}  
                c@{\hspace{0pt}}c@{\hspace{2pt}}  
                c@{\hspace{0pt}}c@{\hspace{2pt}}  
                c@{\hspace{0pt}}c@{\hspace{3pt}}  
                c@{\hspace{0pt}}c@{\hspace{2pt}}  
                c@{\hspace{0pt}}c@{\hspace{2pt}}  
                c@{\hspace{0pt}}c@{\hspace{3pt}}  
                c@{\hspace{0pt}}c@{\hspace{2pt}}  
                c@{\hspace{0pt}}c@{\hspace{2pt}}  
                c@{\hspace{0pt}}}                 

\hline
& & & \multicolumn{3}{c}{\bf{\scshape pl}evel} & & \boldmath$\mathbf\tau$ & & \multicolumn{5}{c}{\bf Baseline} & & \multicolumn{5}{c}{\bf Geometric} & & \multicolumn{5}{c}{\bf{\scshape colts}} \rule{0pt}{2.5ex} \\

\cline{1-2} \cline{4-6} \cline{8-8} \cline{10-14} \cline{16-20} \cline{22-26}

& & & {\footnotesize \bf \#} & & {\footnotesize $\mathbf\imath$\hspace*{12pt}} & & & & \boldmath$\mathbf\eta$ & & \bf{\scshape dacsr} & & \bf{\scshape lcsr} & & \boldmath$\rho$ & & \bf{\scshape dacsr} & & \bf{\scshape lcsr} & & \boldmath$\varrho[0.5]$ & & \bf{\scshape dacsr} & & \bf{\scshape lcsr}\hspace*{-3pt} \rule{0pt}{2.5ex} \\

\cline{1-2} \cline{4-6} \cline{8-8} \cline{10-10} \cline{12-12} \cline{14-14} \cline{16-16} \cline{18-18} \cline{20-20} \cline{22-22} \cline{24-24} \cline{26-26}

\multirow{7}{*}{\begin{sideways}{\bf \textsc{f}rown}\end{sideways}} & \textsc{lapos} & & 18 & & 90,000 & & 1.27 & &   \em 5,000 & &  \em 0.3830 & &  \em 0.0581 & &  1.056 & &  0.3765 & &  0.0876 & & \bf  0.048 & & \bf  0.3478 & & \bf  0.1172\rule{0pt}{2.75ex} \\
& \textsc{m}ax\textsc{e}nt & & 32 & & 160,004 & & 1.70 & &   \em 5,000 & &  \em 0.6275 & &  \em 0.2499 & &  1.031 & &  0.6113 & &  0.2691 & & \bf  0.023 & & \bf  0.5810 & & \bf  0.3326\rule{0pt}{2.25ex} \\
& \textsc{m}orfette & & 20 & & 100,009 & & 1.43 & &   \em 5,000 & &  \em 0.4082 & &  \em 0.0700 & &  1.050 & &  0.3964 & &  0.0979 & & \bf  0.039 & & \bf  0.3850 & & \bf  0.1442\rule{0pt}{2.25ex} \\
& \textsc{mxpost} & & 22 & & 110,017 & & 2.84 & &   \em 5,000 & &  \em 0.7098 & &  \em 0.3621 & &  1.045 & &  0.7012 & &  0.3783 & & \bf  0.039 & & \bf  0.7063 & & \bf  0.4924\rule{0pt}{2.25ex} \\
& \textsc{s}tanford & & 29 & & 145,014 & & 1.91 & &   \em 5,000 & &  \em 0.7250 & &  \em 0.3846 & & \bf  1.034 & & \bf  0.7125 & & \bf  0.3943 & &  0.024 & &  0.6300 & &  0.3905\rule{0pt}{2.25ex} \\
& \textsc{svmt}ool & & 46 & & 230,005 & & 1.41 & &   \em 5,000 & &  \em 0.8518 & &  \em 0.6201 & &  1.022 & &  0.8601 & &  0.6492 & & \bf  0.016 & & \bf  0.8100 & & \bf  0.6522\rule{0pt}{2.25ex} \\
& \textsc{t}n\textsc{t} & & 19 & & 95,018 & & 1.51 & &   \em 5,000 & &  \em 0.4318 & &  \em 0.0829 & & \bf  1.053 & & \bf  0.4188 & & \bf  0.1108 & &  0.038 & &  0.3353 & &  0.1081\rule{0pt}{2.25ex} \\
\cline{1-2} \cline{4-6} \cline{8-8} \cline{10-10} \cline{12-12} \cline{14-14} \cline{16-16} \cline{18-18} \cline{20-20} \cline{22-22} \cline{24-24} \cline{26-26}

\multirow{10}{*}{\begin{sideways}{\bf \textsc{p}enn}\end{sideways}} & fn\textsc{tbl} & & 19 & & 95,007 & & 0.58 & &   \em 5,000 & &  \em 0.3220 & &  \em 0.0346 & &  1.053 & &  0.3226 & &  0.0622 & & \bf  0.058 & & \bf  0.3258 & & \bf  0.1043\rule{0pt}{2.75ex} \\
& \textsc{lapos} & & 13 & & 65,003 & & 0.93 & &   \em 5,000 & &  \em 0.4333 & &  \em 0.0848 & &  1.077 & &  0.4425 & &  0.1258 & & \bf  0.086 & & \bf  0.4193 & & \bf  0.1704\rule{0pt}{2.25ex} \\
& \textsc{m}ax\textsc{e}nt & & 19 & & 95,007 & & 0.60 & &   \em 5,000 & &  \em 0.3393 & &  \em 0.0404 & &  1.053 & &  0.3575 & &  0.0780 & & \bf  0.062 & & \bf  0.3617 & & \bf  0.1289\rule{0pt}{2.25ex} \\
& \textsc{mbt} & & 15 & & 75,035 & & 1.66 & &   \em 5,000 & &  \em 0.4287 & &  \em 0.0817 & &  1.066 & &  0.4344 & &  0.1200 & & \bf  0.054 & & \bf  0.4165 & & \bf  0.1673\rule{0pt}{2.25ex} \\
& \textsc{m}orfette & & 15 & & 75,035 & & 0.52 & &   \em 5,000 & &  \em 0.3410 & &  \em 0.0413 & &  1.066 & &  0.3361 & &  0.0675 & & \bf  0.094 & & \bf  0.3585 & & \bf  0.1260\rule{0pt}{2.25ex} \\
& \textsc{mxpost} & & 17 & & 85,013 & & 1.40 & &   \em 5,000 & &  \em 0.5862 & &  \em 0.2062 & &  1.059 & &  0.5635 & &  0.2215 & & \bf  0.061 & & \bf  0.5124 & & \bf  0.2560\rule{0pt}{2.25ex} \\
& \textsc{s}tanford & & 18 & & 90,031 & & 0.98 & &   \em 5,000 & &  \em 0.5808 & &  \em 0.2003 & &  1.055 & &  0.5841 & &  0.2402 & & \bf  0.050 & & \bf  0.5902 & & \bf  0.3411\rule{0pt}{2.25ex} \\
& \textsc{svmt}ool & & 26 & & 130,008 & & 1.25 & &   \em 5,000 & &  \em 0.7222 & &  \em 0.3807 & &  1.039 & &  0.7115 & &  0.3933 & & \bf  0.028 & & \bf  0.6415 & & \bf  0.4059\rule{0pt}{2.25ex} \\
& \textsc{t}n\textsc{t} & & 12 & & 60,015 & & 0.51 & &   \em 5,000 & &  \em 0.2449 & &  \em 0.0156 & &  1.083 & &  0.2377 & &  0.0320 & & \bf  0.087 & & \bf  0.2131 & & \bf  0.0430\rule{0pt}{2.25ex} \\
& \textsc{t}ree\textsc{t}agger & & 12 & & 60,015 & & 1.32 & &   \em 5,000 & &  \em 0.2667 & &  \em 0.0201 & &  1.083 & &  0.2574 & &  0.0379 & & \bf  0.066 & & \bf  0.2141 & & \bf  0.0432\rule{0pt}{2.25ex} \\
\cline{1-2} \cline{4-6} \cline{8-8} \cline{10-10} \cline{12-12} \cline{14-14} \cline{16-16} \cline{18-18} \cline{20-20} \cline{22-22} \cline{24-24} \cline{26-26}
\hline
\end{tabular}
\end{footnotesize}
\end{center}
\caption{Monitoring of local testing frames with inflation}
\label{table-runs-with-inflation}
\end{table*}

An optimized step function parameter, a common ratio $\rho$ or the
{\sc port} $\varrho[0.5]$ depending on whether the scheduling is
geometric or adaptive, is also included to ensure the credibility of
the results. For the case of the (arithmetic) baselines, as already
said, the common difference $\eta$ matches the tolerance applied
($5*10^3$), which we believe is low enough to guarantee a good
approximation of the real learning curve. All these numerals are
expressed to four decimal digits, using bold (resp. cursive) fonts to
mark the best results among all (resp. the baseline) runs in each
local testing frame.


We discard non-viable local testing frames in ${\mathcal L}$, i.e.
those whose high {\sc pl}evel prevents us from evaluating them with
the observations available, which is why that of {\sc t}ree{\sc
  t}agger on {\sc f}rown is not included in
Table~\ref{table-runs-without-inflation}. Since the purpose of the
collection $\hat{\mathcal L}[1]$ is to illustrate the impact of
unexpected irregularities in the runs of ${\mathcal L}$, we only
include in Table~\ref{table-runs-with-inflation} inflated variants
associated to viable items. The local testing frames for fn{\sc tbl}
and {\sc mbt} on {\sc f}rown are also discarded because their runs
converge just one iteration after the {\sc pl}evel, thus making it
impossible to generate their inflated variants.

\subsubsection{Results on the initial local testing frames}

\begin{figure}[htbp]
\begin{center}
\begin{tabular}{cc}
\hspace*{-.6cm}
\includegraphics[width=0.5\textwidth]{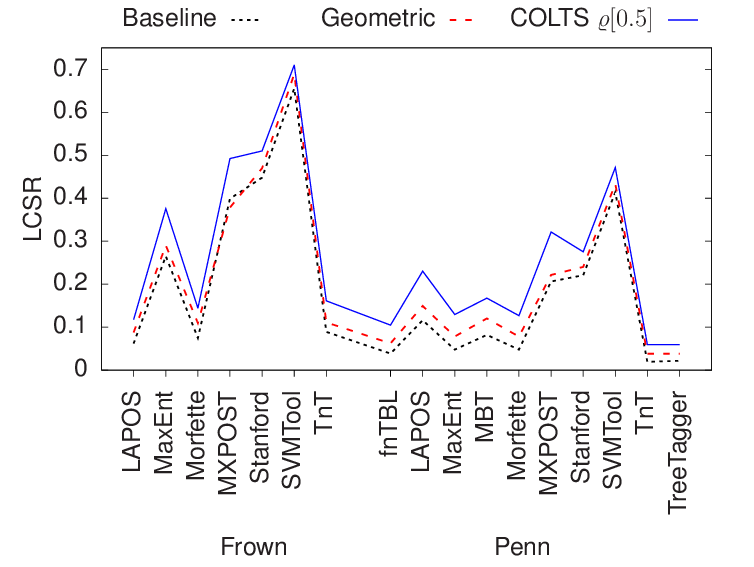} 
&
\includegraphics[width=0.5\textwidth]{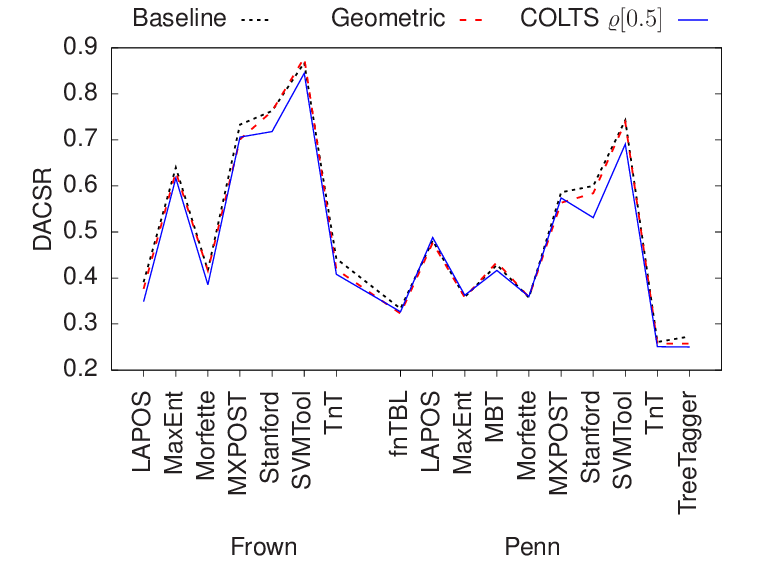}
\end{tabular}
\caption{{\sc lcsr}{\footnotesize s} and {\sc dacsr}{\footnotesize
    s} for runs without inflations.}
\label{fig-LCSR-runs-without-inflations}
\end{center}
\end{figure}

We now focus on the collection ${\mathcal L}$, whose {\sc
  lcsr}{\footnotesize s} (resp. {\sc dacsr}{\footnotesize s}) are
compiled in the left-hand (resp. right-hand) diagram of
Fig.~\ref{fig-LCSR-runs-without-inflations}. The former range from
0.0188 for {\sc t}n{\sc t} on {\sc p}enn using an arithmetic sampling
schedule to 0.7103 for {\sc svmt}ool on {\sc f}rown applying {\sc
  colts}. In percentages, 43.14\% of these values are greater than
0.20, thus proving the adequacy of our parameter setting. Analyzing
each selection approach, this ratio grows to 47.06\% for {\sc colts},
while it drops to 41.18\% for both the geometric and arithmetic
schedules. In more detail, the best performance associates to the
adaptive approach in all local testing frames. Conversely, the
geometric schedule is the second best in 94.12\% of cases, and the
arithmetic one in the remaining 5.88\%. Such prevalence of {\sc colts}
over the geometric selection is mainly due to its increased precision,
while with regard to the arithmetic one it is a consequence of the
lower number of iterations needed to converge. The latter also
explains the better results of the geometric approach against the
arithmetic one. On average, the difference with the baseline is
91.22\% for the adaptive runs (from 1\% to 215.43\% with a standard
deviation of 65.51\%) and 35.42\% for the geometric ones (from 1\% to
101.60\% with a standard deviation of 29.43\%), which gives us an idea
of the magnitude of the flexibility of {\sc colts}.

Regarding {\sc dacsr} values, they range from 0.25 for {\sc t}ree{\sc
  t}agger on {\sc p}enn using adaptive sampling to 0.8788 for {\sc
  svmt}ool on {\sc f}rown with a geometric schedule.  Percentage-wise,
62.75\% of those values are greater than 0.4, rising to 64.71\% for
both arithmetic and geometric runs, while dropping to 58.82\% for the
adaptive ones, once again showing the adequacy of out experimental
setup. Comparing the performances of the three sampling schedules, we
see that arithmetic sampling leads in 70.59\% of all testing frames
and is second best in 23.53\%. Next comes the geometric scheduling
(resp.  {\sc colts}), outperforming the others 11.76\% (resp. 17.65\%)
of the time and being second in 58.82\% (resp. 17.65\%) of tests.

In short and as has been noted above, the better approximation to the
learning curve given by the baseline (arithmetic scheduling) gives it
an advantage in terms of data aquisition costs ({\sc
  dacsr}{\footnotesize s}). Despite this, differences between adaptive
(resp. geometric) and baseline runs are very small, with an average of
4.87\% (resp. 2.15\%), ranging from 0.70\% (resp. 0.08\%) to 11.50\%
(resp. 5.65\%), and a standard deviation of 3.30\%
(resp. 1.81\%). Once model induction ({\sc icsr}) is taken into
account, {\sc colts} is shown to perform with the best overall
learning costs ({\sc lcsr}{\footnotesize s}). However, it is too early
to conclude the superiority of the adaptive scheduling over the rest
of schema compared. A good selection should also help to identify
global rather than local optima, avoiding premature interruptions of
the training. In order to explore this ability, we analyze the {\sc
  lcsr} and {\sc dacsr} metrics on the collection $\hat{\mathcal
  L}[1]$ of inflated variants for ${\mathcal L}$.

\subsubsection{Results on the inflated variants}

\begin{figure}[htbp]
  \begin{center}
\begin{tabular}{cc}
\hspace*{-.6cm}
\includegraphics[width=0.5\textwidth]{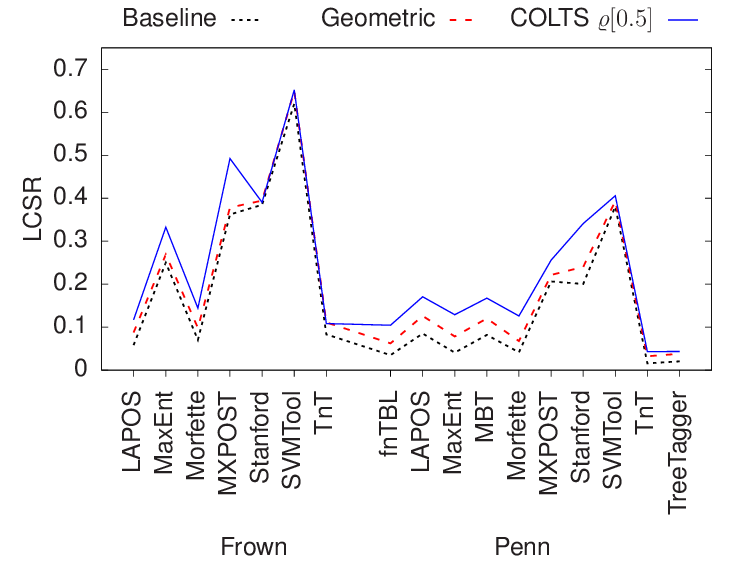} 
&
\includegraphics[width=0.5\textwidth]{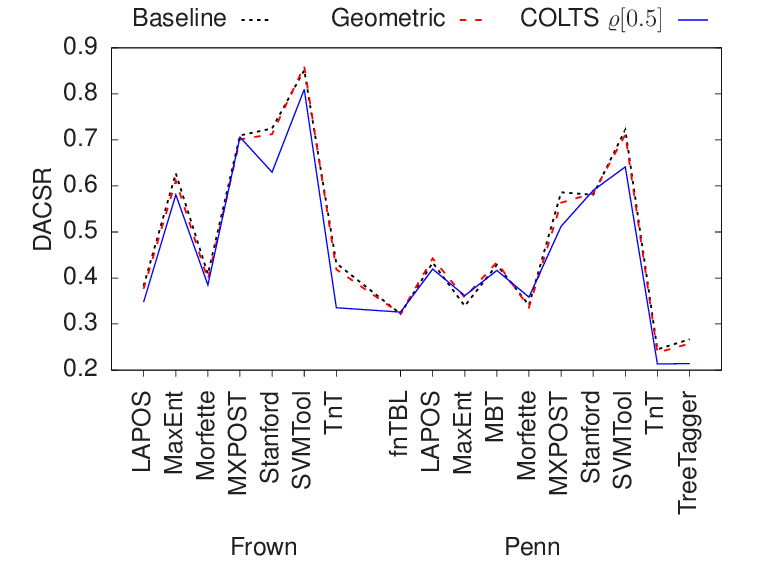}
\end{tabular}
\caption{{\sc lcsr}{\footnotesize s} and {\sc dacsr}{\footnotesize s}
  for runs with inflations.}
\label{fig-LCSR-runs-with-inflations}
\end{center}
\end{figure}

We now focus on the collection $\hat{\mathcal L}[1]$, whose {\sc
  lcsr}{\footnotesize s} (resp. {\sc dacsr}{\footnotesize s}) are
compiled in the left-hand (resp. right-hand) diagram of
Fig.~\ref{fig-LCSR-runs-with-inflations}. The former range from 0.0156
for {\sc t}n{\sc t} on {\sc p}enn with an arithmetic schedule to
0.6522 for {\sc svmt}ool on {\sc f}rown applying {\sc colts}. In
contrast to what happens in ${\mathcal L}$, 41.18\% of the values are
greater than 0.20, which represents a decrease of 1.96\% and reveals
the impact of the irregularities introduced. Looking into each
strategy, results for the arithmetic and geometric ones are the same
as those in ${\mathcal L}$, at 41.18\%. Therefore, the overall
decrease associates to adaptive runs, which drop 5.88\% to
41.18\%. Regarding the best scores, they correspond to our adaptive
approach in 88.24\% of cases, with the remaining 11.76\% favoring the
geometric one. At no time does the arithmetic selection lead the
results. Geometric scheduling obtains the second best {\sc lcsr} in
88.24\% of trials, followed by the adaptive one with 11.76\%. On
average, the difference with the baseline is 72.80\% for the adaptive
runs (from 0.53\% to 172.32\% with a standard deviation of 72.80\%)
and 29.95\% for the geometric ones (from 0.99\% to 76.28\% with a
standard deviation of 25.65\%), while in the arithmetic case it is
9.34\% (from 0\% to 26.83\% with a standard deviation of 6.31\%). All
this further supports the capacity of {\sc colts} to adapt to each
learning process.

Focusing on training resources, {\sc dacsr}{\footnotesize s} range
from 0.2131 for {\sc t}n{\sc t} on {\sc p}enn with an adaptive
schedule to 0.8601 for {\sc svmt}ool on {\sc f}rown with a geometric
one. Of those values, 58.82\% are above 0.40, representing a drop of
3.93\% from the experiments in ${\mathcal L}$, again due to the
irregularities introduced. These losses are not equally shared between
all three strategies. While the percentage of arithmetic runs scoring
above 0.40 remains the same as in ${\mathcal L}$, the one for
geometric (resp. adaptive) runs drops 5.89\% (resp. 5.89\%) to 58.82\%
(resp. 52.93\%). Comparing performances, arithmetic sampling leads in
58.82\% of testing frames and is second best in another 23.53\%, with
geometric (resp. adaptive) sampling besting the others in 17.65\%
(resp. 23.53\%) of cases and running second in another 70.59\%
(resp. 5.88\%). Finally, the difference with the baseline is an
average of 9.81\% for {\sc colts} (from 0.34\% to 24.12\% with an
standard deviation of 7.28\%), 4.38\% for the geometric strategy (from
0.28\% to 8.89\% with an standard deviation of 2.26\%) and 3.30\% for
the arithmetic one (from 0\% to 9.99\% with an standard deviation of
2.33\%).

In summary, as with the initial local testing frames, inflated
arithmetic runs need less training resources ({\sc dacsr}) than the
other ones. But the differences are small enough that, when model
induction effort ({\sc icsr}) is taken into account, {\sc colts} still
comes on top most of the time regarding the overall learning cost
({\sc lcsr}).

\subsubsection{Stability against temporary inflations in performance}

\begin{figure}[htbp]
  \begin{center}
\begin{tabular}{cc}
\hspace*{-.6cm}
\includegraphics[width=0.5\textwidth]{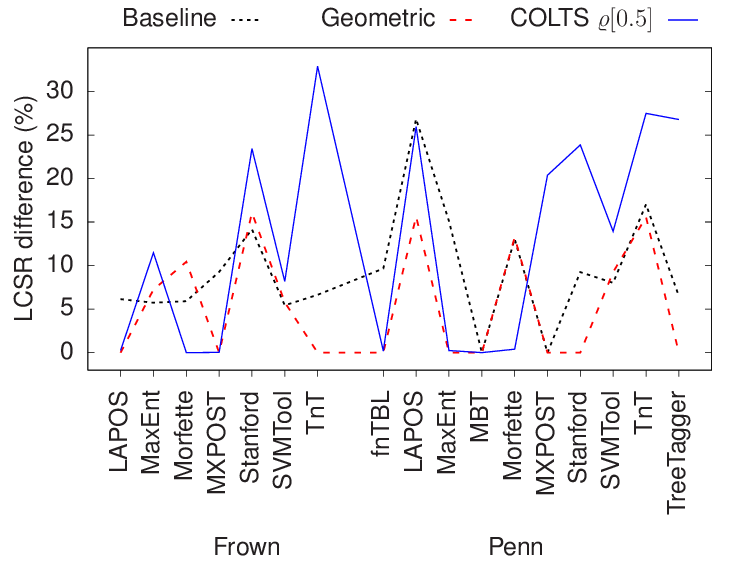} 
&
\includegraphics[width=0.5\textwidth]{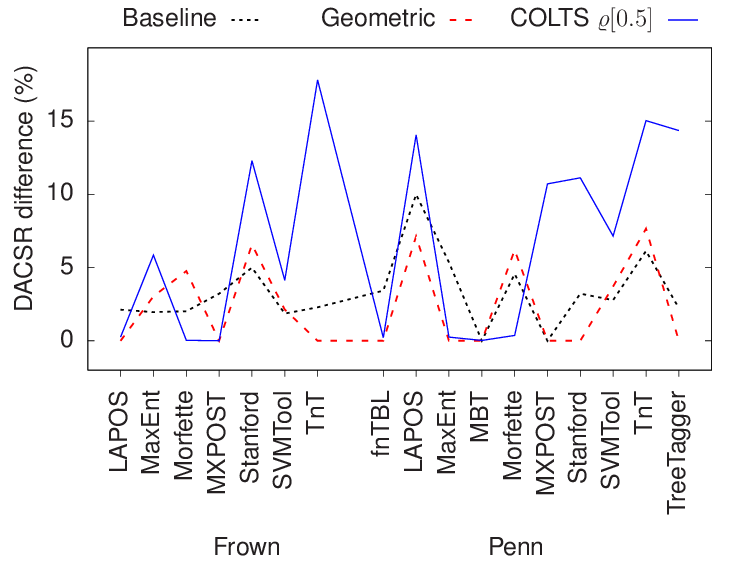}
\end{tabular}
\caption{Differences between {\sc lcsr}{\footnotesize s} (resp. {\sc
    dacsr}{\footnotesize s}) for runs without and with inflations.}
\label{fig-LCSR-runs-different-with-without-inflations}
\end{center}
\end{figure}

The goal is to study the difference between {\sc lcsr}{\footnotesize
  s} (resp. {\sc dacsr}{\footnotesize s}) on each run of the local
testing frames in the collection ${\mathcal L}$ and its corresponding
ones in the associated inflated variant of $\hat{\mathcal L}[1]$, as
shown in the left-hand (resp. right-hand) diagram of
Fig.~\ref{fig-LCSR-runs-different-with-without-inflations}. With
respect to {\sc lcsr} values, geometric scheduling leads the results,
followed by arithmetic sampling with an average difference for the
former (resp. the latter) of 5.48\% (resp. 9.34\%) and a standard
deviation of 6.37\% (resp. 6.31\%). Adaptive scheduling provides us
with the worst average (12.67\%) and standard deviation
(11.94\%). 

Results for {\sc dacsr} are similar. Geometric runs show the best
scores, followed by the arithmetic ones with an average difference of
2.42\% for the former (resp. 3.30\% for the latter) and a standard
deviation of 2.88\% (resp. 2.33\%). {\sc colts} comes last again, with
an average difference of 6.69\% and standard deviation of 6.32\%, but
those differences are even smaller that the ones between {\sc lcsr}
values.

In brief, {\sc colts} shows a similar degree of stability against
inflations to that of the other two selection techniques even, as seen
earlier, its overall learning cost ({\sc lcsr}) is far superior.

\subsubsection{Stability against {\sc port} variations}

\begin{figure}[htbp]
  \begin{center}
\begin{tabular}{cc}
\hspace*{-.6cm}
\includegraphics[width=0.5\textwidth]{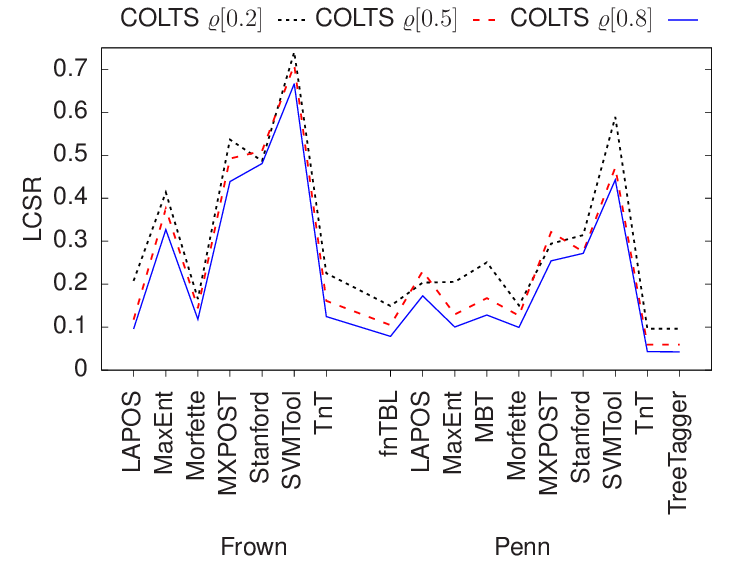}
&
\includegraphics[width=0.5\textwidth]{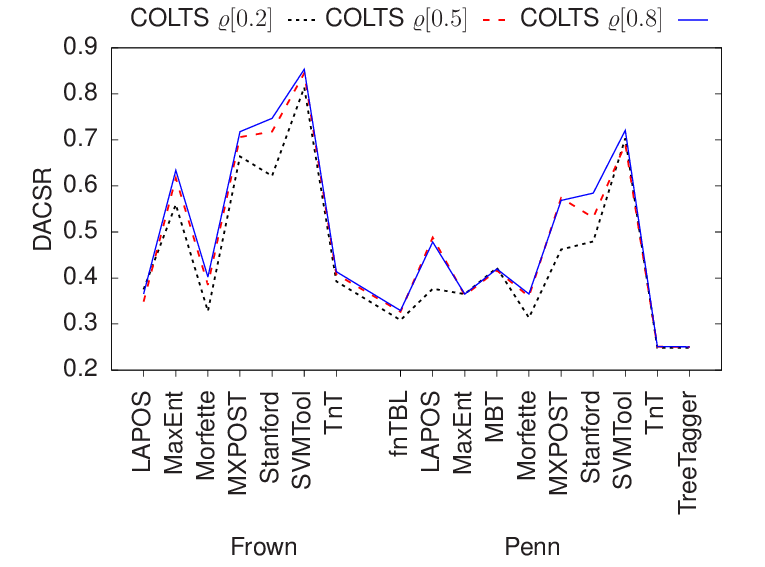}
\end{tabular}
\caption{{\sc lcsr}{\footnotesize s} and {\sc dacsr}{\footnotesize
    s} for adaptive runs without inflations.}
\label{fig-LCSR-adaptive-runs-without-inflations}
\end{center}
\end{figure}

\begin{figure}[htbp]
  \begin{center}
\begin{tabular}{cc}
\hspace*{-.6cm}
\includegraphics[width=0.5\textwidth]{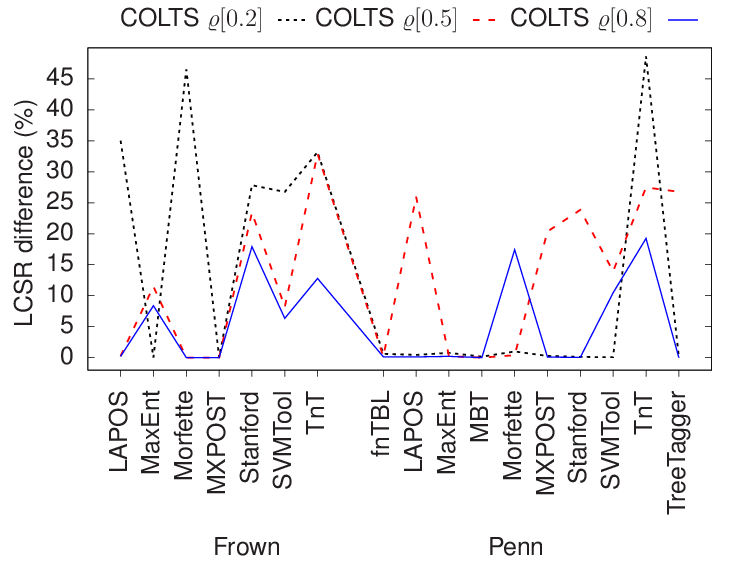}
&
\includegraphics[width=0.5\textwidth]{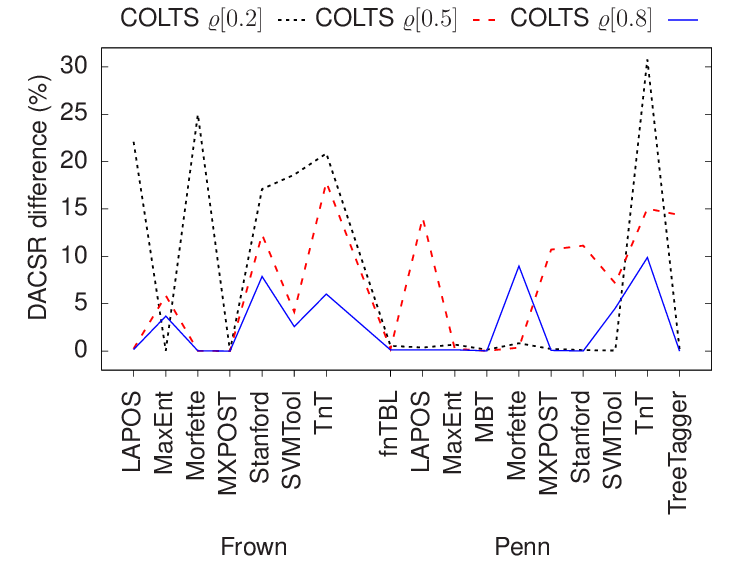}
\end{tabular}
\caption{Differences between {\sc lcsr}{\footnotesize s} (resp. {\sc
    dacsr}{\footnotesize s}) for adaptive runs without and with
  inflations.}
\label{fig-LCSR-adaptive-runs-different-with-without-inflations}
\end{center}
\end{figure}

Although the above tests support the idea that {\sc colts} performs
better than its opponents, they were obtained from a particular {\sc
  port} parameter $\varrho[0.5]$. The question that remains is whether
such a conclusion can be generalized, which involves surveying the
evolution of performance regarding the {\sc port} chosen. We then
extend the collection of step functions associated to each local
testing frame with new adaptive schedules corresponding to {\sc port}
parameters $\varrho[0.2]$ and $\varrho[0.8]$, thereby increasing the
number of runs using this kind of scheduling to three. As $\psi \in
(0,1]$, $\varrho[0.5]$ is an intermediate value for $\varrho[\psi]$
  while the other two are extreme ones, this provides a representative
  comparative framework on adaptive selection. Focusing on these runs,
  we now review each local testing frame ${\mathcal
    L}[\mathcal{A}^\pi[\mathcal{D}^{\mathcal{K}}_{\Xi_\eta}],\wp_{\eta},\tau]
  \in {\mathcal L}$ to study their variations with respect to {\sc
    lcsr} (resp. {\sc dacsr}), as shown in the left-hand
  (resp. right-hand) diagram of
  Fig.~\ref{fig-LCSR-adaptive-runs-without-inflations}. We also
  compare them with their corresponding inflated variants in
  $\hat{\mathcal L}[1][\hat{\mathcal
      A}^\pi[\mathcal{D}^{\mathcal{K}}_{\Xi_\eta}],\wp_{\eta},\tau]
  \in \hat{\mathcal L}[1]$, as seen in
  Fig.~\ref{fig-LCSR-adaptive-runs-different-with-without-inflations}.

  The new experiments show that {\sc lcsr} seems to be inversely
  proportional to the value $\psi$ of $\varrho[\psi]$, reaching the
  highest (resp. smallest) scores with $\varrho[0.2]$
  (resp. $\varrho[0.8]$). The average {\sc lcsr} for the former
  (resp. the latter) in ${\mathcal L}$ is 0.3015 (resp. 0.2285) with a
  standard deviation of 0.1820 (resp. 0.1777), while these rates are
  0.2621 and 0.1838 for $\varrho[0.5]$, illustrating the reliability
  of the adaptive technique irrespective of the {\sc port}
  used. Regarding stability against irregularities in the working
  hypotheses, the average difference between the {\sc lcsr} for runs
  in ${\mathcal L}$ using $\varrho[0.2]$ (resp. $\varrho[0.8]$) as
  {\sc port} and their inflated variants in $\hat{\mathcal L}[1]$ is
  13.04\% (resp. 5.49\%), with a standard deviation of 17.92\%
  (resp. 7.14\%). For $\varrho[0.5]$ the percentages are 12.67\% and
  11.94\% respectively.

  Regarding training resources ({\sc dacsr}), model induction savings
  ({\sc icsr}) with low $\psi$ values translate into lower efficiency
  in approximating the real learning curve. Thus, contrary to what has
  just been shown for {\sc lcsr}{\footnotesize s}, {\sc dacsr} and
  $\psi$ are directly correlated. So, runs using $\varrho[0.2]$
  (resp. $\varrho[0.8]$) have the worst (resp.  best) average {\sc
    dacsr} of 0.4519 (resp. 0.4980) with a standard deviation of
  0.1613 (resp. 0.1789), while runs using $\varrho[0.5]$ represent a
  middle ground with an average {\sc dacsr} of 0.4870 and standard
  deviation of 0.1728. It is worth noticing that the differences
  between those averages are very small, underlying once again the
  stability of {\sc colts} with respect to {\sc port}. Regarding the
  effect of irregularities in the learning curve, the average
  difference between the {\sc dacsr} for runs in ${\mathcal L}$ using
  $\varrho[0.2]$ (resp. $\varrho[0.8]$) as {\sc port} and their
  inflated variants in $\hat{\mathcal L}[1]$ is 8.10\% (resp. 2.60\%),
  with a standard deviation of 10.90\% (resp. 3.45\%). For
  $\varrho[0.5]$ these percentages are 6.69\% and 6.32\% respectively,
  matching the behaviour previously observed for {\sc lcsr}.

  Briefly, even the performance is inversely (resp. directly)
  proportional to the parameter $\psi$ of $\varrho[\psi]$ with respect
  to overall learning (resp. data acquisition) costs, just as expected
  and reflected by the {\sc lcsr} (resp. {\sc dacsr}) metric, the
  adaptive scheduling ({\sc colts}) always maintains acceptable
  values.

\section{Conclusions}
\label{section-conclusions}

We develop an adaptive scheduling ({\sc colts}) for non-active
adaptive sampling in order to reduce the training effort in the
generation of {\sc ml}-based {\sc pos} taggers. Formally, the
technique demonstrates its correctness with respect to its working
hypotheses. Namely, it provides the minimal spacing needed between
consecutive instances for ensuring that the next observation is
relevant in learning terms. Based on a geometrical criterion, the
selection task is modeled from a sequence of learning trends which
iteratively approximates the learning curve. In every cycle, the
algorithm calculates the distance to the next case as the minimal one
from where we are sure that the hypothesis of concavity can no longer
be guaranteed. The selection schedule described can also be configured
according to a {\sc port} parameter controlling the interaction
between the speed of learning and the size of samples. Regarding
robustness, our analysis determines that the factors at play are
similar to those affecting the stability of the halting
condition. Hence, we entrust its treatment to the mechanisms then
applied, which first entails having a tool to avoid potentially
intractable distortions.

With a view to allow its practical value to be beyond doubt, a
demanding and competitive testing framework has been designed for our
proposal. Given a learner, a halting condition and a training data
base, we categorize schedules according to their performance, seen in
terms of both overall learning costs and training resources needed to
generate a model with a given level of accuracy. The normalization of
the experimental conditions, including a formally correct proximity
criterion to measure and stabilize such convergence, ensures that the
standards of evidence do not favour any particular option, thus
guarantying their reliability.

The results corroborate the expectations established in the
theoretical framework as well as its stability. That way, since it
does not depend on domain-specific requirements, the doors are open to
exploit {\sc colts} for reducing the workload in {\sc ml} uses other
than the case study considered. The issue is then particularly
relevant to tasks in the {\sc nlp} sphere, where the learning
procedures are more and more challenging in a variety of applications
such as machine translation, text classification or parsing.

\section*{Acknowledgements}
\begin{small}
  Research partially funded by the Spanish Ministry of Economy and
  Competitiveness through projects TIN2017-85160-C2-1-R and
  TIN2017-85160-C2-2-R, and by the Galician Regional Government under
  projects ED431C 2018/50 and ED431D 2017/12.
\end{small}

\begin{footnotesize}

\begin{thebibliography}{10}

\bibitem{AounallahQuirionMineau04}
Mohamed Aounallah, S\'ebastien Quirion, and Guy~W. Mineau.
\newblock Distributed data mining vs. sampling techniques: A comparison.
\newblock In {\em Advances in Artificial Intelligence}, pages 454--460.
  Springer-Verlag, 2004.

\bibitem{Attenberg:2011:ILD:1964897.1964906}
Josh Attenberg and Foster Provost.
\newblock Inactive learning? {D}ifficulties employing active learning in
  practice.
\newblock {\em ACM SIGKDD Explorations Newsletter}, 12(2):36--41, 2011.

\bibitem{Biemann:2006:UPT:1557856.1557859}
Chris Biemann.
\newblock Unsupervised part-of-speech tagging employing efficient graph
  clustering.
\newblock In {\em Proceedings of the 21st International Conference on
  Computational Linguistics and 44th Annual Meeting of the Association for
  Computational Linguistics: Student Research Workshop}, pages 7--12, Sydney,
  2006.

\bibitem{Bloodgood:2009:MSA:1596374.1596384}
Michael Bloodgood and K.~Vijay-Shanker.
\newblock A method for stopping active learning based on stabilizing
  predictions and the need for user-adjustable stopping.
\newblock In {\em Proceedings of the 13th Conference on Computational Natural
  Language Learning}, pages 39--47, Boulder, 2009.

\bibitem{Branch1999}
Mary~Ann Branch, Thomas~F. Coleman, and Yuying Li.
\newblock A subspace, interior, and conjugate gradient method for large-scale
  bound-constrained minimization problems.
\newblock {\em SIAM Journal on Scientific Computing}, 21(1):1--23, 1999.

\bibitem{Brants2000}
Thorsten Brants.
\newblock {TnT}: A statistical part-of-speech tagger.
\newblock In {\em Proceedings of the 6th Conference on Applied Natural Language
  Processing}, pages 224--231, Seattle, 2000.

\bibitem{Brill1995a}
Eric Brill.
\newblock Transformation-based error-driven learning and natural language
  processing: A case study in part-of-speech tagging.
\newblock {\em Computational Linguistics}, 21(4):543--565, 1995.

\bibitem{CaseBerg:01}
George Casella and Roger Berger.
\newblock {\em Statistical Inference}.
\newblock {Duxbury Resource Center}, 2001.

\bibitem{Castro18}
Francisco~M. Castro, Manuel~J Mar{\'i}n-Jim{\'e}nez, Nicol{\'a}s Guil, Cordelia
  Schmid, and Karteek Alahari.
\newblock {End-to-End Incremental Learning}.
\newblock In Vittorio Ferrari, Martial Hebert, Cristian Sminchisescu, and Yair
  Weiss, editors, {\em {ECCV 2018 - European Conference on Computer Vision}},
  volume 11216 of {\em Lecture Notes in Computer Science}, pages 241--257,
  Munich, Germany, 2018. {Springer}.

\bibitem{Chen:2013:PNA:2568488.2568754}
Jianhua Chen.
\newblock Properties of a new adaptive sampling method with applications to
  scalable learning.
\newblock In {\em Proceedings of the 2013 IEEE/WIC/ACM International Joint
  Conferences on Web Intelligence (WI) and Intelligent Agent Technologies (IAT)
  - Volume 01}, WI-IAT '13, pages 9--15, Washington, DC, USA, 2013. IEEE
  Computer Society.

\bibitem{Chernoff52}
H.~Chernoff.
\newblock A measure of asymptotic efficiency for tests of a hypothesis based on
  the sums of observations.
\newblock {\em Annals of Mathematical Statistics}, 23:409--507, 1952.

\bibitem{Chrupala2008}
Gzregorz Chrupala, Georgiana Dinu, and Josef van Genabith.
\newblock Learning morphology with {M}orfette.
\newblock In {\em Proceedings of the 6th International Conference on Language
  Resources and Evaluation}, pages 2362--2367, Marrakech, 2008.

\bibitem{Clark2010}
Alexander Clark, Chris Fox, and Shalom Lappin.
\newblock {\em The {H}andbook of {C}omputational {L}inguistics and {N}atural
  {L}anguage {P}rocessing}.
\newblock John Wiley \& Sons, Hoboken, 2010.

\bibitem{Cohn:1994:IGA:189256.189489}
David Cohn, Les Atlas, and Richard Ladner.
\newblock Improving generalization with active learning.
\newblock {\em Machine Learning}, 15(2):201--221, 1994.

\bibitem{Collins2002}
Michael Collins.
\newblock Discriminative training methods for hidden {M}arkov models: theory
  and experiments with perceptron algorithms.
\newblock In {\em Proceedings of the 2002 Conference on Empirical Methods in
  Natural Language Processing (Vol. 10)}, pages 1--8, Philadelphia, 2002.

\bibitem{Daelemans1996a}
Walter Daelemans, Jakub Zavrel, Peter Berck, and Steven Gillis.
\newblock {MBT}: A memory--based part-of-speech tagger generator.
\newblock In {\em Proceedings of the 4th Workshop on Very Large Corpora}, pages
  14--27, Copenhagen, 1996.

\bibitem{Domingo:2002:ASM:593433.593526}
Carlos Domingo, Ricard Gavald\`{a}, and Osamu Watanabe.
\newblock Adaptive sampling methods for scaling up knowledge discovery
  algorithms.
\newblock {\em Data Mining and Knowledge Discovery}, 6(2):131--152, 2002.

\bibitem{French99}
Robert~M. French.
\newblock Catastrophic forgetting in connectionist networks.
\newblock {\em Trends in Cognitive Sciences}, 3:128--135, 1999.

\bibitem{Freund:1997:SSU:263100.263123}
Yoav Freund, H.~Sebastian Seung, Eli Shamir, and Naftali Tishby.
\newblock Selective sampling using the query by committee algorithm.
\newblock {\em Machine Learning}, 28(2-3):133--168, September 1997.

\bibitem{FreyFisher99}
L.J. Frey and D.H. Fischer.
\newblock Modeling decision tree performance with the power law.
\newblock In {\em Proceedings of the 7th International Workshop on Artificial
  Intelligence and Statistics}, pages 59--65, Fort Lauderdale, 1999.

\bibitem{Furnkranz98}
Johannes F\"urnkranz.
\newblock Integrative windowing.
\newblock {\em Journal of Artificial Intelligence Research}, 8:129--164, 1998.

\bibitem{Giesbrecht2009}
E.~Giesbrecht and S.~Evert.
\newblock Is part-of-speech tagging a solved task? {A}n evaluation of {POS}
  taggers for the {G}erman web as corpus.
\newblock In {\em Proceedings of the 5th Web as Corpus Workshop}, pages 27--35,
  San Sebastian, 2009.

\bibitem{Gimenez2004}
Jes\'us Gim\'enez and Llu\'{\i}s M\'arquez.
\newblock {SVMT}ool: A general {POS} tagger generator based on support vector
  machines.
\newblock In {\em Proceedings of the 4th International Conference on Language
  Resources and Evaluation}, pages 43--46, Lisbon, 2004.

\bibitem{Hinrichs2010}
Lars Hinrichs, Nicholas Smith, and Birgit Waibel.
\newblock Manual of information for the part-of-speech-tagged, post-edited
  '{B}rown' corpora.
\newblock {\em ICAME Journal}, 34:189--233, 2010.

\bibitem{Hoeffding63}
Wassily Hoeffding.
\newblock Probability inequalities for sums of bounded random variables.
\newblock {\em Journal of the American Statistical Association},
  58(301):13--30, 1963.

\bibitem{Howard66}
Ronald~A. Howard.
\newblock Decision analysis: Applied decision theory.
\newblock In {\em Proceedings of the 4th International Conference on
  Operational Research}, pages 55--71, Cambridge, 1966.

\bibitem{John96staticversus}
George John and Pat Langley.
\newblock Static versus dynamic sampling for data mining.
\newblock In {\em Proceedings of the 2nd International Conference on Knowledge
  Discovery and Data Mining}, pages 367--370, Portland, 1996.

\bibitem{KadiePhd}
C.M. Kadie.
\newblock {\em Seer: Maximum Likelihood Regression for Learning-Speed Curves}.
\newblock {PhD} thesis, Univ. of Illinois at Urbana-Champaign, 1995.

\bibitem{Kapoor05}
Aloak Kapoor and Russell Greiner.
\newblock Learning and classifying under hard budgets.
\newblock In {\em Machine Learning: ECML 2005}, pages 170--181.
  Springer-Verlag, 2005.

\bibitem{Last:2009:IDM:1557019.1557076}
Mark Last.
\newblock Improving data mining utility with projective sampling.
\newblock In {\em Proceedings of the 15th ACM SIGKDD International Conference
  on Knowledge Discovery and Data Mining}, pages 487--496, Paris, 2009.

\bibitem{Leite:2012:SCA:2358856.2358868}
Rui Leite, Pavel Brazdil, and Joaquin Vanschoren.
\newblock Selecting classification algorithms with active testing.
\newblock In {\em Proceedings of the 8th International Conference on Machine
  Learning and Data Mining in Pattern Recognition}, pages 117--131, Berlin,
  2012.

\bibitem{Lewis:1994:SAT:188490.188495}
David~D. Lewis and William~A. Gale.
\newblock A sequential algorithm for training text classifiers.
\newblock In {\em Proceedings of the 17th Annual International ACM SIGIR
  Conference on Research and Development in Information Retrieval}, pages
  3--12, Dublin, 1994.

\bibitem{Lipton1993}
Richard~J. Lipton, Jeffrey~F. Naughton, Donovan~A. Schneider, and S.~Seshadri.
\newblock Efficient sampling strategies for relational database operations.
\newblock {\em Theoretical Computer Science}, 116(1):195 -- 226, 1993.

\bibitem{Losing18}
Viktor Losing, Barbara Hammer, and Heiko Wersing.
\newblock Incremental on-line learning: A review and comparison of state of the
  art algorithms.
\newblock {\em Neurocomputing}, 275:1261--1274, 2018.

\bibitem{Lynch2003}
James~F. Lynch.
\newblock Analysis and application of adaptive sampling.
\newblock {\em Journal of Computer and System Sciences}, 66(1):2 -- 19, 2003.
\newblock Special Issue on \{PODS\} 2000.

\bibitem{Mair2007}
Christian Mair and Geoffrey Leech.
\newblock The {Freiburg-Brown} corpus ('{F}rown') ({POS-t}agged version).
\newblock Albert-Ludwigs-Universit\"at Freiburg and University of Lancaster.
  Available for academic use through {http://clu.uni.no/icame/}, 2007.

\bibitem{Marcus1999}
Mitchell~P. Marcus, Beatrice Santorini, Mary~Ann Marcinkiewicz, and Ann Taylor.
\newblock Treebank-3 {LDC99T42}.
\newblock Web download file. Linguistic Data Consortium, Philadelphia, 1999.

\bibitem{Meek:2002:LSM:944790.944798}
Christopher Meek, Bo~Thiesson, and David Heckerman.
\newblock The learning-curve sampling method applied to model-based clustering.
\newblock {\em The Journal of Machine Learning Research}, 2:397--418, March
  2002.

\bibitem{Ngai2001}
Grace Ngai and Radu Florian.
\newblock Transformation-based learning in the fast lane.
\newblock In {\em Proceedings of the 2nd Meeting of the North American chapter
  of the Association for Computational Linguistics on Language technologies},
  pages 1--8, Pittsburgh, 2001.

\bibitem{Provost:1999:EPS:312129.312188}
Foster Provost, David Jensen, and Tim Oates.
\newblock Efficient progressive sampling.
\newblock In {\em Proceedings of the 5th ACM SIGKDD International Conference on
  Knowledge Discovery and Data Mining}, pages 23--32, San Diego, 1999.

\bibitem{Quinlan83}
John~Ross Quinlan.
\newblock Learning efficient classification procedures and their application to
  chess end games.
\newblock In {\em Machine Learning. An Artificial Intelligence Approach},
  volume~1, pages 463--482, 1983.

\bibitem{Ratnaparki1996}
Adwait Ratnaparkhi.
\newblock A maximum entropy model for part-of-speech tagging.
\newblock In {\em Proceedings of the 1996 Conference on Empirical Methods in
  Natural Language Processing}, pages 133--142, Philadelphia, 1996.

\bibitem{Reichart:2010:TLC:1870568.1870579}
Roi Reichart, Omri Abend, and Ari Rappoport.
\newblock Type level clustering evaluation: New measures and a pos induction
  case study.
\newblock In {\em Proceedings of the Fourteenth Conference on Computational
  Natural Language Learning}, CoNLL '10, pages 77--87, Stroudsburg, PA, USA,
  2010. Association for Computational Linguistics.

\bibitem{Saar-Tsechansky04}
Maytal Saar-Tsechansky and Foster Provost.
\newblock Active sampling for class probability estimation and ranking.
\newblock {\em Machine Learning}, 54(2):153--178, 2004.

\bibitem{Schmid1994}
Helmut Schmid.
\newblock Probabilistic part-of-speech tagging using decision trees.
\newblock In {\em Proceedings of the International Conference on New Methods in
  Language Processing}, pages 44--49, Manchester, 1994.

\bibitem{Schmid:2008:ECP:1599081.1599179}
Helmut Schmid and Florian Laws.
\newblock Estimation of conditional probabilities with decision trees and an
  application to fine-grained pos tagging.
\newblock In {\em Proceedings of the 22Nd International Conference on
  Computational Linguistics - Volume 1}, COLING '08, pages 777--784,
  Stroudsburg, PA, USA, 2008. Association for Computational Linguistics.

\bibitem{Schutze:2006:PTP:1183614.1183709}
Hinrich Sch\"{u}tze, Emre Velipasaoglu, and Jan~O. Pedersen.
\newblock Performance thresholding in practical text classification.
\newblock In {\em Proceedings of the 15th ACM International Conference on
  Information and Knowledge Management}, pages 662--671, Arlington, 2006.

\bibitem{Seung:1992:QC:130385.130417}
H.~S. Seung, M.~Opper, and H.~Sompolinsky.
\newblock Query by committee.
\newblock In {\em Proceedings of the 5th Annual Workshop on Computational
  Learning Theory}, pages 287--294, Pittsburgh, 1992.

\bibitem{Song:2012:CSP:2390524.2390661}
Hyung-Je Song, Jeong-Woo Son, Tae-Gil Noh, Seong-Bae Park, and Sang-Jo Lee.
\newblock A cost sensitive part-of-speech tagging: Differentiating serious
  errors from minor errors.
\newblock In {\em Proceedings of the 50th Annual Meeting of the Association for
  Computational Linguistics: Long Papers (Vol. 1)}, pages 1025--1034, Jeju
  Island, 2012.

\bibitem{KATRIN08.335}
Katrin Tomanek and Udo Hahn.
\newblock Approximating learning curves for active-learning-driven annotation.
\newblock In {\em Proceedings of the 6th International Conference on Language
  Resources and Evaluation}, pages 1319--1324, Marrakech, 2008.

\bibitem{Toutanova2003}
Kristina Toutanova, Dan Klein, Christopher~D. Manning, and Yoram Singer.
\newblock Feature-rich part-of-speech tagging with a cyclic dependency network.
\newblock In {\em Proceedings of the 2003 Annual Conference of the North
  American chapter of the Association for Computational Linguistics on Human
  Language Technology (Vol. 1)}, pages 173--180, Edmonton, 2003.

\bibitem{Tsai14}
Cheng-Hao Tsai, Chieh-Yen Lin, and Chih-Jen Lin.
\newblock Incremental and decremental training for linear classification.
\newblock {\em Proceedings of the ACM SIGKDD International Conference on
  Knowledge Discovery and Data Mining}, pages 343--352, 2014.

\bibitem{Tsuruoka2011}
Yoshimasa Tsuruoka, Yusuke Miyao, and Jun'ichi Kazama.
\newblock Learning with lookahead: Can history-based models rival globally
  optimized models?
\newblock In {\em Proceedings of the 15th Conference on Computational Natural
  Language Learning}, pages 238--246, Portland, 2011.

\bibitem{VanHalteren1999}
Hans van Halteren.
\newblock Performance of taggers.
\newblock In {\em Syntactic Wordclass Tagging}, pages 81--94. Kluwer Academic
  Pub., Hingham, 1999.

\bibitem{VilaresDarribaRibadas16}
Manuel Vilares, V\'{\i}ctor~M. Darriba, and Francisco~J. Ribadas.
\newblock Modeling of learning curves with applications to {POS} tagging.
\newblock {\em Computer Speech \& Language}, 41:1--28, 2017.

\bibitem{Vlachos:2008:SCA:1349893.1350099}
Andreas Vlachos.
\newblock A stopping criterion for active learning.
\newblock {\em Computer Speech and Language}, 22(3):295--312, 2008.

\bibitem{Watanabe:2005:SST:1162426.1162428}
Osamu Watanabe.
\newblock Sequential sampling techniques for algorithmic learning theory.
\newblock {\em Theoretical Computer Science}, 348(1):3--14, 2005.

\bibitem{WeissTian08}
GaryM. Weiss and Ye~Tian.
\newblock Maximizing classifier utility when there are data acquisition and
  modeling costs.
\newblock {\em Data Mining and Knowledge Discovery}, 17(2):253--282, 2008.

\bibitem{Winston75}
P.~H. Winston.
\newblock Learning structural descriptions from examples.
\newblock In {\em The Psychology of Computer Vision}, pages 157--209.
  McGraw-Hill, New York, 1975.

\end{thebibliography}

\end{footnotesize}

\end{document}